# Fundamentals of *Physical AI*

Vahid Salehi

Institute of Engineering Design of Mechatronic Systems
`salehi-d@hm.edu`
DOI: 10.71015/z6mc6967

*I would like to dedicate this work to the **University of California, Berkeley**,
where I had the privilege to serve as a Visiting Scholar.*

## Abstract

This work will elaborate the fundamental principles of physical artificial intelligence (*Physical AI*) from a scientific and systemic perspective. The aim is to create a theoretical foundation that describes the physical embodiment, sensory perception, ability to act, learning processes, and context sensitivity of intelligent systems within a coherent framework. While classical AI approaches rely on symbolic processing and data driven models, *Physical AI* understands intelligence as an emergent phenomenon of real interaction between body, environment, and experience. The six fundamentals presented here are embodiment, sensory perception, motor action, learning, autonomy, and context sensitivity, and form the conceptual basis for designing and evaluating physically intelligent systems. Theoretically, it is shown that these six principles do not represent loose functional modules but rather act as a closed control loop in which energy, information, control, and context are in constant interaction. This circular interaction enables a system to generate meaning not from databases, but from physical experience, a paradigm shift that understands intelligence as an physical embodied, material process. *Physical AI* understands learning not as parameter adjustment, but as a change in the structural coupling between agents and the environment. To illustrate this, the theoretical model is explained using a practical scenario: An adaptive assistant robot supports patients in a rehabilitation clinic. Thanks to its soft, sensor supported joints, it can physically sense how much weight a person can bear. It learns from resistance, movement, and body language to precisely regulate its strength and recognizes when it needs to actively intervene or hold back. This example illustrates that physical intelligence does not arise from abstract calculation, but from immediate, embodied experience. It shows how the six fundamentals interact in a real system: embodiment as a prerequisite, perception as input, movement as expression, learning as adaptation, autonomy as regulation, and context as orientation. In general, this work is intended to contribute to the theoretical and methodological foundation of a new generation of intelligent systems that act physically, learn, and act responsibly in the world. The principles formulated here are intended to serve as a guide for researchers and engineers to further develop *Physical AI* from an abstract idea into a practical design concept.

**Keywords:** Embodied AI, Artificial Intelligence, Physical AI, Fundamental Physics, Fundamental Robotics, Robotic, Autonomous Driving, Intelligent Health Engineering, AI Engineering, System of Systems Engineering, Cyberphysical Systems, Digital Twin, Virtual Engineering, Computer Aided Engineering, Internet of Things (IoT)

# 1   Introduction

The development of artificial intelligence (AI) has reached a level of momentum in recent decades that has encompassed all scientific and social fields. From data analysis and medical diagnostics to autonomous systems, artificial intelligence has enabled a new level of machine perception and decision making. Nevertheless, one central limitation remains: the physical dimension of intelligence. Although classical AI systems excel at recognizing patterns and making decisions in the digital realm, they lack the ability to act, sense, and learn in the physical world. This is precisely where the concept of physical artificial intelligence (*Physical AI*) comes in. It understands intelligence not as abstract computing power but as an physically embodied, situational, and energy coupled ability to act meaningfully in a physical environment. The basic theoretical idea of *Physical AI* goes beyond the familiar separation of software and hardware. Consider cognitive processes as emergent properties of physical interaction. In contrast to purely symbolic or data driven models, meaning arises here from movement, resistance, contact, and context. The body of a system becomes the carrier and mediator of its intelligence. This shifts the focus from representation to action, from model to experience. This view is based on the work of Brooks [30], Clark [28], and Varela et al. [27], who argued as early as the 1990s that thinking cannot be meaningfully or completely described without the body. This paper builds on this theoretical perspective and develops a conceptual framework that describes the fundamental structural elements of physical intelligence. *Physical AI* is understood here as a closed, recursive system in which energy, information, control, and context are inextricably linked. This system approach, which is based on the principles of system theory and cybernetics of Ashby [33], emphasizes the reciprocal coupling of perception and action. In such a system, cognitive states arise not through data accumulation, but through dynamic interactions between sensory, motor, and environmental systems. To make the theoretical considerations tangible, *Physical AI* can be illustrated using a practical example: An adaptive rehabilitation robot supports patients during their motor recovery. This robot has a flexible sensor supported body structure that allows it to sense pressure, movement, and muscle tension. It learns not from simulations or data sets, but from physical contact with humans through resistance, balance, and feedback. With each interaction, its behavior is fine tuned until it can anticipate movements without causing pain or disrupting the healing process. This example clearly shows that physical intelligence does not arise at the data level, but in the exchange between materiality, perception, and adaptation. *Physical AI* therefore understands learning as structural adaptation, not as merely statistical training. It is about acquiring skills through direct experience in the real world. The rehabilitation robot develops its "experience" by modulating force and movement, which in turn changes its control models. This feedback loop of learning by doing is at the heart of the physically intelligent principle. In this perspective, cognitive architecture merges with physical reality, creating systems that not only calculate but also "feel" and "understand". The significance of this approach extends far beyond the field of robotics. *Physical AI* could be used in the future in industry, healthcare, or space travel anywhere where systems need to interact physically, react autonomously, and collaborate safely with humans. In nursing, for example, physically intelligent robots could help support movement sequences, reduce muscle strain, and document therapeutic progress. In manufacturing, on the other hand, they could adapt to material tolerances without being reprogrammed. This work is therefore intended not only as a theoretical contribution but also as a methodological foundation to guide future research on embodiment, sensors, action competence, and context sensitivity of physical intelligence. The following chapters define and analyze six fundamental principles of *Physical AI*, meaning embodiment, sensory perception, motor competence, learning ability, autonomy, and context sensitivity, and place them in a closed system model. Together, these six principles form a recursive, energy coupled network that forms the basis of any future physically intelligent architecture. Overall, this work is intended as an invitation to rethink the idea of artificial intelligence: not as disembodied computation, but as a living system in physical resonance with its environment. The rehabilitation robot serves as a practical example of the transition from abstract intelligence to tangible, acting, and learning intelligence, symbolizing that thinking and body, perception and ethics, physics, and consciousness are not separate but deeply interwoven.

## 2 Literature Survey

In the early stages of development, the focus was on reflecting on the black box problem and the question of the explainability of AI systems. In 2018, Adadi and Berrada [1] laid the methodological foundation for the subsequent discussion on transparent, explainable physical systems with their overview of Explainable Artificial Intelligence (XAI). They identified trust and traceability as key epistemic conditions for the social acceptance of AI. At the same time, Jin et al. [2] investigated the fundamentals of mobile intelligence in robotics and emphasized the role of sensory fusion, navigation models, and deep learning structures for autonomous systems. These early works already show the transition from purely data driven AI to systems that integrate physical perception and movement into their learning model. From 2019 on wards, applications combining physical interaction, sensor data, and control logic will increasingly come to the fore. Radanliev et al. [3] examined the challenges of artificial intelligence in cyber physical systems (CPS) and designed a hierarchical framework model for AI supported decision making processes. In parallel, Zhou et al. [4] addressed the emerging field of "edge intelligence", i.e., the shift of computing power to the edge of the network a crucial prerequisite for real time responses in physical environments. In their overview of collaborative robotics, Emeoha Ogenyi et al. [5] emphasized the importance of sensors and actuators for human machine interaction. Their work marks the transition from an algorithmic to a physically interactive perspective. Finally, Mehrabi et al. [6] pointed out the ethical and social dimensions of *Physical AI* applications in their study on bias and fairness in AI a topic that was systematically explored in greater depth in subsequent years. Overall, this phase shows that *Physical AI* arises from the tension between technical efficiency and social responsibility. The foundations of the later embodiment thesis are laid here, initially through explainability, fairness, and physical interfaces. The year 2020 marks the beginning of the actual theoretical establishment of *Physical AI*. Miriyev and Kova [7] were the first to call for an integrative education and research agenda on physical artificial intelligence, understood as an interdisciplinary field between robotics, materials science, and AI. In this phase, biological analogies, self organization, and energetic control models come into focus. With their work on physics guided machine learning, Willard et al. [8] provided a methodological basis for hybrid models that combine empirical data with physical laws. This connection, from statistics to mechanics, forms the epistemological backbone of *Physical AI*. The year 2021 is considered a turning point. Sitti [9] coined the term physical intelligence (PI) and distinguished between computational intelligence in the "brain" and physical intelligence in the "body" of the agent. This was the first clear formulation of the idea of the duality of cognitive and material intelligence. Duan et al. [10] expanded this perspective with a comprehensive encyclopedia of physical embodied AI and evaluated nine simulation platforms and three central research tasks: visual exploration, navigation, and embodied questioning. These works show that embodiment and environmental interaction are no longer marginal topics but have become central research axes of physical embodied AI. In parallel, Wagner [11] developed a conceptual framework for automating literature reviews using AI meta approach that shows how machine systems can reproduce scientific thinking themselves. This phase establishes physical intelligence as an independent paradigm: intelligence is understood as an emergent property of physical, energetically coupled systems. The years 2022 and 2023 are characterized by methodological deepening and disciplinary integration. Duan et al. [12] systematized machine learning approaches to modeling "intuitive physics" inspired by cognitive studies on human perception of physical relationships. This work highlighted that machine learning, that explicitly integrates physical principles produces more robust and generalizable models. Hao et al. [13] comple-mented this perspective with an overview of physics informed machine learning (PI-ML). They showed that the integration of prior physical knowledge into neural architectures is still insufficiently developed and advocated for models that embed physical constraints directly into their structure. In 2023, the focus shifted from methodological foundations to applications and evaluation systems. Melnik et al. [14] provided an overview of benchmarks for physical reasoning, an attempt to measure the "physical reasoning ability" of AI. De la Torre López et al. [15] analyzed the use of AI in automated literature analysis and described the role of humans as a reflexive control instance, a concept that was later taken up in human in the loop intelligence. Mahligawati et al. [16] expanded the discourse to include the educational aspect by examining the role of AI in physics education, thereby advancing the transfer of physical intelligence into didactic contexts. Finally, Chang et al. [17] highlighted security critical aspects, in particular AI based attacks and countermeasures at the physical communication level a topic that later became central to the discussion of autonomous systems and ethical control. This phase shows that *Physical AI* is now establishing itself not only as a theoretical ideal, but as a technical and methodological reality. The discourse is shifting from "if" to "how" from the existence of physical intelligence to the design of its architecture. In 2024 and 2025, research reaches a new depth. *Physical AI* is understood here for the first time as a closed theoretical and empirical framework. Yang Liu et al. [18] positioned physical embodied AI as a crucial stepping stone on the path to artificial general intelligence (AGI). They analyzed four core areas embodied perception, interaction, agent architectures, and sim to real transfer and emphasized that multimodal world models form the basis of future cognitive systems. Jiao et al. [19] expanded this perspective to include the AI physics connection, i.e., the influence of physical disciplines on machine learning. Their approach resolves the dichotomy of physics as a model and AI as a tool, viewing both as mutually inspiring systems. Siam et al. [20] explored the field of Artificial Intelligence of Things (IoT) a key discipline for Distributed *Physical AI*. They showed how sensory perception, communication, and adaptive control converge into a convergent *Physical AI* infrastructure. Conceptual completion will take place in 2025. Liu et al. [21] presented one of the first systematic classifications of physics based generation models in their overview of Generative *Physical AI* in Vision. They emphasize that future models must take physical plausibility into account as well as visual quality an indication that physical realism understanding is becoming a central evaluation category. In parallel, Dewi et al. [22] define *Physical AI* as an independent field of research, distinguish between Integrated and Distributed PAI, and classify the discipline between materials science, computer science, chemistry, and biology. Their work is considered the most comprehensive systematization of the field to date. Xiang et al. [23] deepen the theoretical foundation by defining physics based reasoning as a basic cognitive principle and formulating "world models" as a prerequisite for explainable, predictive intelligence. Long et al. [24] supplement this with their overview of embodied learning in physical simulators. They show that the combination of external simulation and internal world modeling is the key to sim to real transfer. Finally, Bousetouane et al. [25] present the Physical Retrieval Augmented Generation (Ph-RAG) pattern an architecture that links physical perception, cognition, and action into a modular whole. Their work combines *Physical AI* with large language models (LLMs), marking the transition to industrial implementation. The latest research is rounded off by Chen et al. [26], who present acoustic world models based on physical wave information, opening up a previously little noticed dimension of physical perception: acoustics as an epistemic carrier of material properties. This wealth of work shows that *Physical AI* 2025 has become a mature, interdisciplinary field of research based equally on systematic theory, empirical methodology, and ethical reflection. The Cosmos World Foundation Model Platform for *Physical AI* represents a key advance in this context, as it introduces so called World Foundation Models as universally applicable models that can be fine tuned for various *Physical AI* applications [40]. This lays the foundation for the development of digital twins that can significantly improve the training of *Physical AI* systems. Similarly, Chen et al. examine the development from perceptual to behavioral intelligence in the context of physical embodied AI and divide the field of research into three core areas: perception, decision making, and execution, with particular emphasis on the influence of large scale foundation models [41]. In addition, Maraita et al. provide a comprehensive overview of human robot interaction, focusing on AI driven physical interaction systems. In particular, haptic feedback, safety aspects, and learning from demonstrations are identified as key factors for future *Physical AI* systems [42]. Another central research topic is the security and robustness of *Physical AI* systems. Xing et al. analyze vulnerabilities and attack scenarios in physical embodied AI systems and classify threats according to both exogenous causes, such as physical attacks and cyber security risks and endogenous factors, such as sensor errors or software defects [43]. Wang and Wei complement this with comprehensive overviews of physical adversarial attacks in the field of computer vision and highlight the security challenges that arise when AI systems come into direct contact with the physical world [44], [45]. Nguyen focuses specifically on security related vulnerabilities in surveillance systems and shows that physical adversarial attacks in real world *Physical AI* applications represent a previously underestimated risk [46]. In addition to security research, numerous studies are devoted to the application of *Physical AI* in specialized domains. Sumner et al. provide a systematic overview of the use of AI in physical rehabilitation and highlight in particular the benefits of robotic devices, gaming systems, and wearable sensors for physical therapy [47]. Progress is also being made in the field of elderly care: Wang et al. describe intelligent physical service robots that are specifically designed to meet the needs of older people, thereby enabling a new form of personalized support [48]. Farajtabar et al., on the other hand, focus on contact based human robot interactions and discuss ways to achieve safer physical collaboration between humans and machines [49]. At the same time, physics based AI approaches are gaining in importance. Meng et al. investigate how physical priors can be integrated into the generation of 3D and 4D content to ensure structural consistency and realistic motion dynamics [50]. Banerjee et al. systematize the research field of physics informed reinforcement learning (PIRL) by proposing a new taxonomy for integrating physical laws into learning processes [51]. Finally, Angelis et al. show that AI based symbolic regression is increasingly capable of reconstructing physical laws and equations directly from data a milestone on the road to knowledge discovering AI [52]. The technological infrastructure of *Physical AI* is also evolving dynamically. Singh and Sipola investigate the role of edge AI hardware and software for *Physical AI* systems that rely on real time processing at the periphery [53], [54]. The work of Baccour et al. points in a similar direction, analyzing pervasive AI for the Internet of Things (IoT) and highlighting resource efficient, distributed computing architectures as a key factor for the scalability of *Physical AI* applications [55]. Another research focus concerns the evaluation and benchmarking of *Physical AI* systems. Wong et al. investigate robotic navigation and manipulation using physical simulators and analyze their properties and hardware requirements for physical embodied AI research [56]. Collins et al. have already provided an important basis for this by presenting a systematic overview of physical simulators for robotic applications, thereby creating a foundation for the development of standardized evaluation methods [57]. Finally, an interdisciplinary trend can be observed that links physics, engineering, and computer science more closely. Alhousseini et al. provide insight from a physical perspective into how neural networks are able to reconstruct physical theories without explicit prior knowledge [58]. In addition, Ye et al. show that *Physical AI* approaches are increasingly being applied in the field of wireless communication at the physical layer [59]. Several key findings emerge from this expanded literature analysis. First, the physical security of AI systems remains a critical factor, especially in safety critical applications. Second, significant progress is evident in health related applications, particularly in rehabilitation and elderly care. Third, the importance of edge computing as a necessary infrastructural basis for real time processes in *Physical AI* is emphasized. Fourth, recent developments illustrate that AI can not only apply physical laws, but also discover them independently. Finally, secure and efficient physical interaction between humans and machines is at the center of future developments. Despite these advances, significant research gaps remain. For example, there is currently a lack of established ethical and security related frameworks for *Physical AI*, while aspects such as energy efficiency and sustainability of *Physical AI* systems are insufficiently addressed. Furthermore, there is no uniform standardization of evaluation metrics, and interdisciplinary cooperation models have only been developed to a limited extent. Further research works carried out in [6067] are considering the holistic approach of systems development. These gaps mark key starting points for future research projects in the field of *Physical AI*. Based on the literature survey following actions are required:

1. Early research on explainable AI focused on the traceability of models without asking how these systems actually experience reality. The first need for action was therefore to understand intelligence no longer as abstract computation but as embodied, energetically embedded activity. Only a material body enables the interaction with the environment that anchors knowledge in a situation and generates meaning. Our solution approach in this work will be the integration of FUNDAMENTAL 1 for Physical embodiment. The need for world reference.

2. In previous AI research, sensory systems were primarily regarded as measuring instruments that collect data without interpreting it. The central need for action was to understand perception as an active process of meaning formation, in which every sensory experience is interpreted in a context of action. Only this recursive coupling of perception and movement turns raw signals into an understanding relationship with the environment. Our solution approach in this work will be the integration of FUNDAMENTAL 2 for Sensory perception from signal to semantics.

3. In many robotic architectures, action remained a consequence of the calculation, not its source. The need for action therefore lies in understanding action as an epistemic act as an experiment with which a system tests hypotheses about its environment. Intelligence does not arise from observation, but from intervention: a physically acting agent learns because it feels the consequences of its actions. Our solution approach in this work will be the integration of FUNDAMENTAL 3 for Motor action competence knowledge through doing.

4. The data driven paradigm of AI proved fragile in the face of changing environments and incomplete data. The identified need for action was to base learning on physical feedback: systems must learn from experience, not from statistics. Only when perception, action, and consequence are in a closed loop can behavior develop in a stable and context sensitive manner. Our solution approach in this work will be the integration of FUNDAMENTAL 4 for the learning ability experience as a source of adaptation.

5. With increasing autonomy of the system, it became clear that freedom of choice without responsibility is dangerous. The need for action was to define autonomy not as the absence of control, but as self determination within comprehensible limits. An intelligent system must be able to explain and limit its actions, not because it is restricted, but because reflexivity is part of its intelligence. Our solution approach in this work will be the integration of FUNDAMENTAL 5 for autonomy regulated self control instead of blind freedom.

6. Many existing models treat context as additional information, not as a defining structural element. The resulting need for action was to understand context as a dynamic system variable that continuously modulates perception, action, and learning. A context sensitive system not only understands what to do, but also recognizes when and why an action is appropriate. Our solution approach in this work will be the integration of FUNDAMENTAL 6 for Context sensitivity Situational understanding instead of static rules

Finally, the literature revealed a fragmentation between physical modeling, cognitive architecture, and ethical reflection. The overarching need for action was to combine these areas in a coherent framework in which the physical, cognitive, and normative dimensions support each other. This is precisely where the six fundamentals come in: they define the minimal but sufficient conditions under which a system can act intelligently in the world.

## 3  Fundamentals of *Physical AI*

Physical intelligence arises where perception, movement, energy, and meaning are inextricably interrelated. The concept of physical artificial intelligence (*Physical AI*) is based on the fundamental assumption that thinking does not take place in a vacuum, but in a body it is a material, energy consuming, vulnerable body that interacts with its environment and learns from this interaction [10]. No matter how sophisticated, the computing architecture remains incomplete as long as it has no access to the physical world. While classical AI systems function as disembodied entities computing cores that operate on abstract symbols or data, *Physical AI* describes intelligence as an physically embodied process that unfolds in the interplay of perception and action. This understanding is correlated with the early work of Brooks [11], Clark [12] and Varela et al. [27], who showed that intelligence is not an isolated internal activity but arises from the coupling of the brain, body and environment. Ashby [14] supplemented this view with his cybernetic principle of self regulation: systems must learn to maintain their internal states while exchanging energy and information with their environment. *Physical AI* understands this self regulation as the basis of learning. According to this view, all intelligent behavior is a process of feedback in a loop of perception, movement, adaptation, and control. These processes are not linear, but circular: what a system does changes what it perceives, and this changed perception in turn influences what it does. Intelligence therefore does not arise through central planning but through a continuous negotiation of stability in the flow of energy and information. Figure 1 shows this relationship as a closed control loop: energy, control, and context circulate in an endless process of coordination. Within this circle, six fundamentals can be identified that together form the structural basis of physical intelligence.

1. Embodiment: the physical basis of experience and action
2. Sensory perception: the conversion of energy into meaning,
3. Motor Action competence: the ability to adapt movements to the environment and the goal,
4. Learning Ability: the ongoing change of internal structures through experience,
5. Autonomy: independent regulation of goals and actions within safe limits,
6. Context Sensitivity: the embedding of actions in social, spatial, and emotional contexts.

**The Definition of Physical AI therefore is:** *Physical AI refers to artificial intelligence systems that not only process information and make decisions in the digital space but also interact directly with the physical world and they use sensors to perceive their environment, model dynamic states of the real world, make autonomous decisions on how to act, and implement these decisions using actuators (e.g. robotic arms, driving movements, gripping devices). The aim is to achieve continuous adaptation and learning ability under conditions of uncertainty, limited perception, and physical laws.*

These six principles are not modules that operate separately from one another, but rather aspects of a living system. An agent that physically acts intelligently senses, learns, and decides in a single, intertwined process. To make theoretical considerations tangible, an example will serve as a common thread: an adaptive rehabilitation robot that supports patients after neurological or orthopedic injuries. This robot has a soft sensor supported body architecture and is able to sense force, balance, and movement. It learns to adjust its support to muscle resistance, irregular breathing patterns, and the rhythm of therapy. If the patient loses balance, the robot reacts immediately, balancing forces and gently stabilizing. This interaction is not merely an "application" of intelligence; it is intelligence: a continuous balancing act between physical resonance and cognitive control. The six fundamentals of *Physical AI* are explained in detail in the following. Each section begins with a definition, explains the theoretical significance, and describes how the principle is embodied in the rehabilitation robot. Figure 1 illustrates the six important FUNDAMENTALS of *Physical AI*.

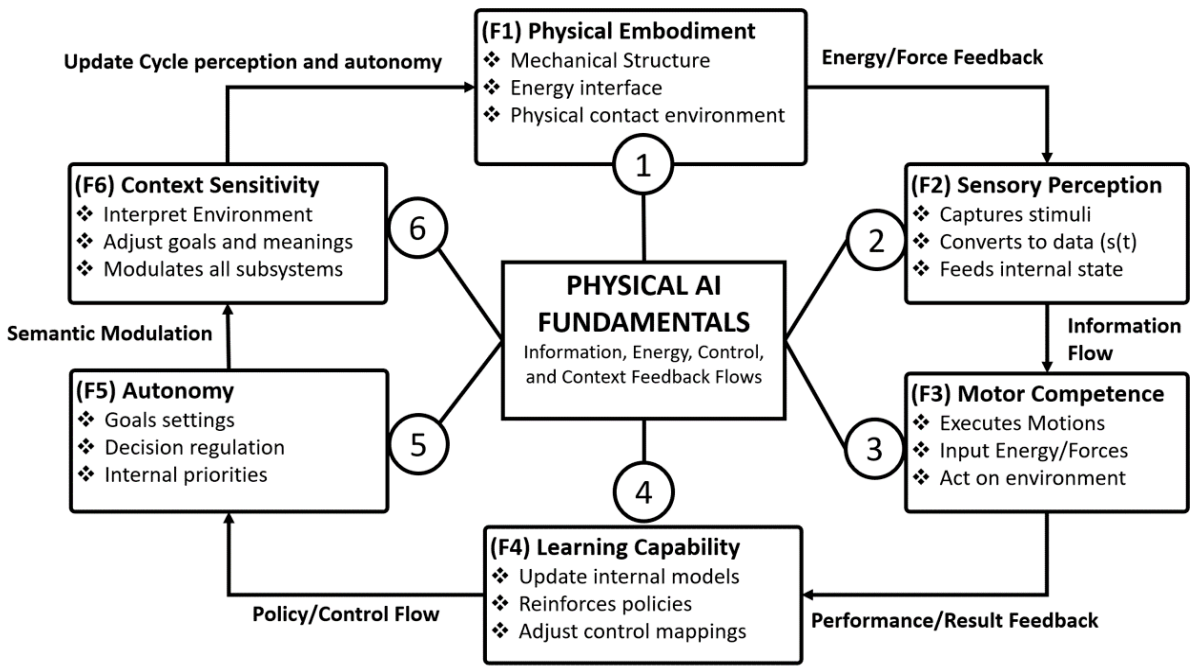

Figure 1: Six important FUNDAMENTALS of *Physical AI*

## 3.1 Fundamental 1 is the Physical Embodiment

***Definition 1 (Embodiment).*** *Physical Embodiment refers to the physical embodiment of cognitive processes in a material body that serves as a sensory motor medium and generates meaning through physical interactions.*

Figure 2 entitled (F1) Physical Embodiment describes how the body forms the basis for perception, action, and learning. The point is that the body is not just a physical object, but the central medium through which cognition is possible in the first place. Determines what a system can perceive, what it can respond to, and what it can ultimately learn. The upper part lists three fundamental areas that form the physical foundation: structure, energy and forces, and material properties. Structure includes, for example, joints and the framework of a body, in other words, everything that determines form and mobility. The area of energy and forces refers to balance and load, i.e., how the body absorbs and distributes forces. Finally, material properties describe how flexible or inert the body is, in other words, how strongly it can adapt to or resist movements. These three aspects have a direct impact on the functional systems of the body. The structure gives rise to the sensory system, which receives information about the body's own condition and the environment. The distribution of energy and forces is closely related to the actuator system, i.e., the motor system that enables movement. The material properties, in turn, influence energy control, i.e., the way energy is used and regulated in the body. At the bottom of the diagram is the interaction with the environment, the point at which all these systems interact. This interaction includes contact, movement, and resistance. It describes how the body relates to its environment, how it senses it, reacts to it, and adapts to it. In general, the diagram makes it clear that physical characteristics and cognitive abilities are inextricably linked.

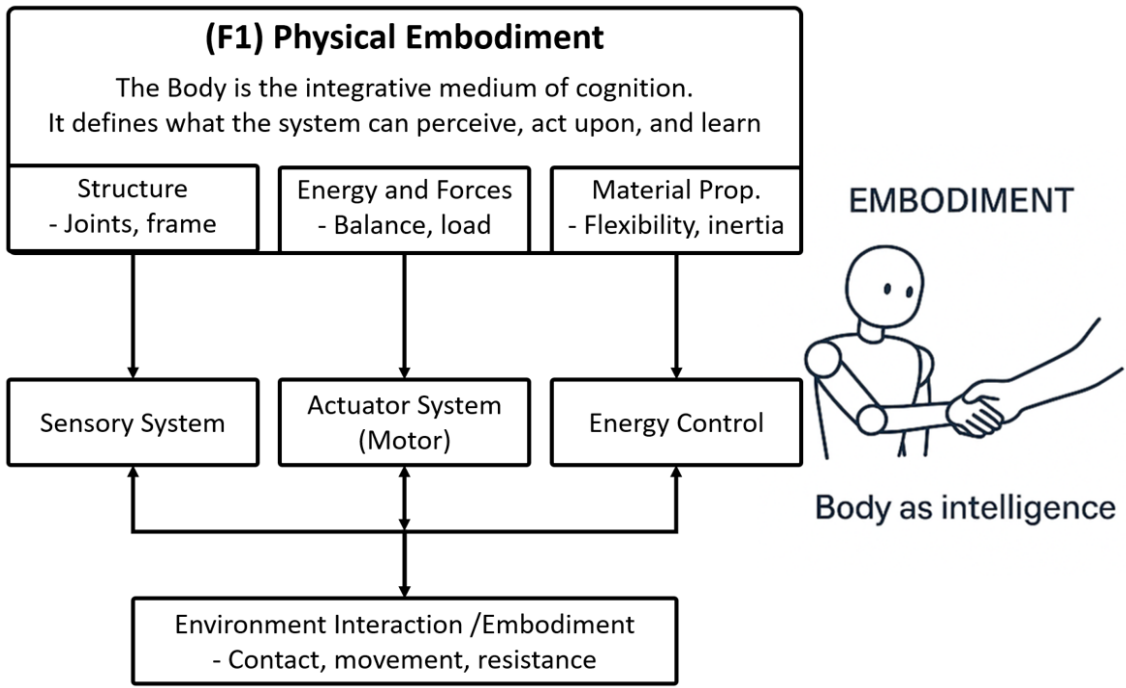

Figure 2: Fundamental 1 is the Physical Embodiment

The body is not merely a tool controlled by the mind; it is itself an essential part of thinking, learning, and acting. Embodiment is the foundation of any form of intelligence that is to exist in the real world. Without a body, there is no perception; without perception, there is no experience; and without experience, there is no meaning. Intelligence, as Brooks [11] put it, arises from the interaction of a body with its environment, not from abstract symbol manipulation. This insight has radically changed the idea of the "thinking head": the body is not an auxiliary organ of the mind but its origin. From the perspective of systems theory of Ashby [14], embodiment can be understood as the medium through which a system absorbs, transforms, and controls energy. A body mediates between the environment and the internal organization. Every movement, every sensory stimulus is part of an energetic cycle in which information is created. The materiality of the body thus becomes a condition for cognitive processes; the mechanics of grasping, the elasticity of a joint, or the inertia of a movement become computational operations themselves. Pfeifer [12] describes this as morphological computation:

the body participates in cognitive work by pre-processing information through its form and material behavior. This principle is immediately apparent in the example of the rehabilitation robot. Its body consists of soft, flexible segments that conform to the human arm without fixing it in place. Every touch creates physical communication: pressure sensors detect force distribution, elastic joints store energy and relieve the patient. If the person moves too quickly or too far, the robot senses the change in resistance and automatically dampens the movement. This "feeling in the body" happens before any digital model intervenes. The robot's body calculates mechanically, not symbolically. Embodiment therefore also means that intelligence is distributed and decentralized. It does not reside in a central processor, but in the interaction of many local dynamics: micro stretching, friction, vibration, balance. The intelligence of the rehabilitation robot arises from this diversity of micro physical feedback . When it guides a patient, each interaction is a small dialog of energy exchange and balance search. This can be formally described as a feedback system:

$$I(t) = f(E(t), S(t), A(t), C(t)), \tag{1}$$

Here, $I(t)$ stands for the current state of intelligence, $E(t)$ for energy flows, $S(t)$ for sensory states, $A(t)$ for actions and $C(t)$ for context. These variables influence each other mutually. Embodiment is the physical infrastructure that enables this flow. The practical result is remarkable: the rehabilitation robot "knows" when to leave because its structure reacts to the slightest changes. When the patient regains strength, the robot withdraws minimally. It does not have to "calculate" this decision; it results from the interaction between the body and the environment. This is precisely where the elegance of physical intelligence lies: it arises from things themselves, not above them. Embodiment is thus more than a design principle; it is an ontological statement. The body is the condition of possibility for intelligence. In a physical system such as the rehabilitation robot, mechanics, perception, and meaning merge into a unified process. The robot thinks by feeling. And it feels because it is built to resonate with the world.

### 3.2 Fundamental 2 is the Sensory Perception

***Definition 2 (Sensory Perception).*** *Sensory perception refers to the continuous process by which a physical system translates energy, movement, and structure in its environment into meaningful internal states. It is not a passive measurement, but an active process of coupling and resonance between the system and its environment.*

Figure 3 entitled (F2) Sensory Perception describes how perception converts physical stimuli into internal states. It shows how a system be it a biological organism or an artificial entity perceives its environment through various senses, processes this information, and derives knowledge or actions from it. It starts with the input, i.e., the input signals from the environment. These include light (e.g.,through a camera), sound (through a microphone), and touch (through pressure sensors). These stimuli form the basis for all perception processes. This is followed by the processing step. Here, the incoming information is analyzed and prepared. Feature extraction takes place, in which relevant characteristics are filtered out of the raw data. At the same time, noise reduction is performed to make perception clearer and more reliable. Finally, context weighting is performed, which means that certain information is rated as more or less important depending on the situation. This processed information flows into the perceptual model. This model creates a state representation, i.e., a kind of internal image or model of the current state of the environment and one's own body. Uncertainty also plays a role here, the system takes into account that perception is never completely certain and always contains room for interpretation. From here, the information is transferred to cognition and learning. In this step, the perceived states are linked to existing experiences and knowledge. The system learns to recognize patterns, understand connections, and better plan future actions. Finally, this entire process leads to action (actuator/motor, and action). This means that perception and cognition are translated into concrete movements or reactions.

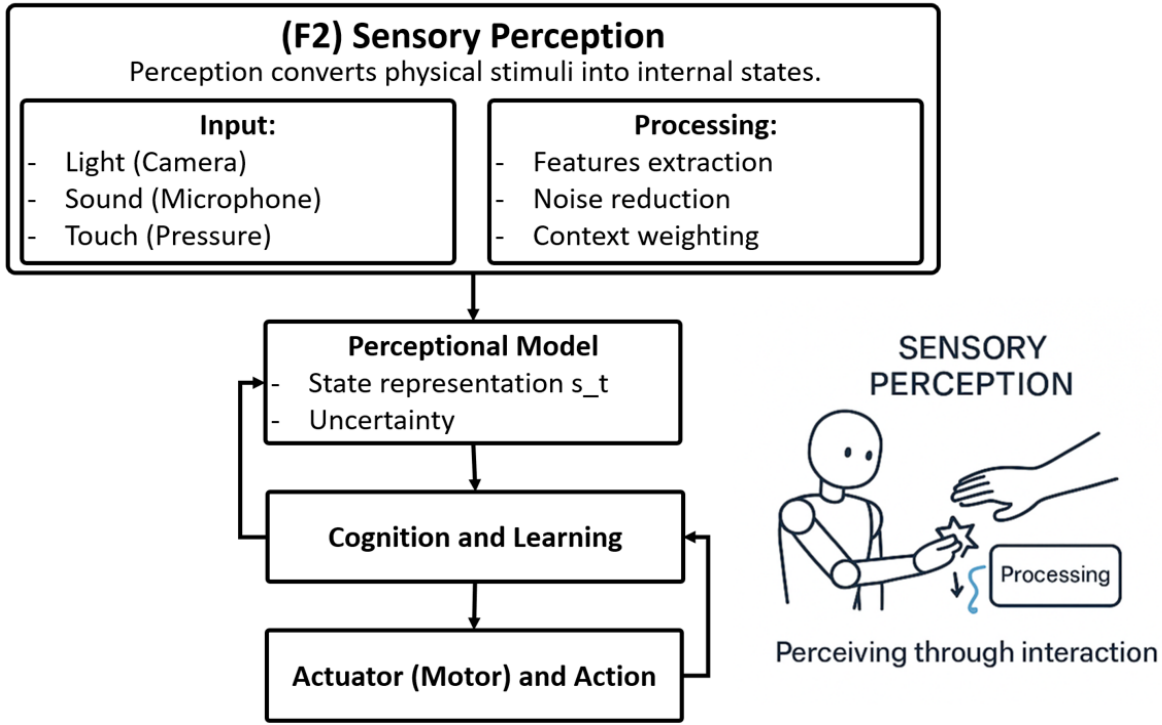

Figure 3: FUNDAMENTAL 2 Sensory Perception

At the same time, the results of these actions flow back into perception, and the system constantly checks how its actions change the environment and adapts accordingly. Overall, the diagram shows that perception is an active and dynamic process. It begins with the reception of physical stimuli, leads through complex processing steps to an internal representation of the world, and finally results in actions that, in turn, trigger new perceptions. This creates a continuous cycle of feeling, thinking, and acting. Perception is the gateway to the world. Without it, all intelligence would remain closed to itself, unable to respond to changes in its environment. In classic AI systems, perception is often a matter of data collection and classification: sensors provide input that is interpreted by algorithms. But for *Physical AI*, perception means more than just reading signals; it is a mutual scan, an interplay of energy and meaning [15]. The basic assumption is that a system does not simply record what "is there" but actively constitutes the world through its sensors and its body. Every sensor arrangement is an interpretive structure, not a neutral channel. The physics of perception pressure, light, temperature, sound is filtered through material properties and translated into the logic of the system. This means that perception is always embodied, never objective. Ashby [16] has already spoken of the principle of requisite variety: a system can only reflect the same complexity of the environment as it has its own internal states. Sensory intelligence therefore consists of having sufficient variance in one's own body to mirror the variance of the world. The rehabilitation robot embodies this in an impressive way. It has a wide range of sensors: tactile sensors on gripping surfaces, strain sensors on joints, microphones for detecting acoustic signals, and optical sensors for recognizing movement paths. When it guides a patient's arm, a sensory conversation takes place between human and machine. A sudden resistance, a change in muscle tone, or a quiet exhalation are not analyzed separately, but interpreted in context: perhaps the movement was too fast, perhaps fatigue is setting in. The robot immediately adjusts the flow of force, not by calculation, but by sensory accompaniment. This type of perception is fundamentally active. The robot "feels" the movement and changes it at the same time. This creates a dynamic feedback loop in which perception and action become one. In theory, this is referred to as active perception by Varela et al. [27]: a system creates the world it perceives through its interactions with it. This view contrasts with classical cognitive psychology, which understands perception as representation. In *Physical AI*, perception is a process of coordination. The system is not informed by the world; it attunes itself to it. Multi modality is a particularly important aspect here. The more sensors communicate with each other, the richer the internal model of the environment becomes. The rehabilitation robot integrates tactile and acoustic input to develop a kind of "somatic awareness." When assisting a patient, it hears the rhythm of their breathing and simultaneously senses the movement of their muscles. The combination of these signals creates an intuitive idea of the person's state, whether tense, relaxed, or uncertain. This form of sensory integration does not make the robot human, but it brings it closer to the meaning of human sensitivity. The decisive theoretical moment lies in the generation of meaning from energy. In *Physical AI*, every sensory input is understood as a transformation of energy into information. Perception is therefore an energetic semantic process. The robot does not simply feel force; it "signifies" it: too strong a pull on the arm implies not only more energy, but also a higher risk of injury a semantic assignment that causes it to reduce force. This creates a primitive form of meaning evaluation from which responsibility can follow. Perception in *Physical AI* can therefore be formalized as a circular process:

$$S(t+1) = f_S(S(t), A(t), E(t), C(t)), \qquad (2)$$

where $S(t)$ describes the sensory state at time $t$, $A(t)$ describes the current actions, $E(t)$ describes the energy flows, and $C(t)$ describes the context. Sensory perception is not linear, but reciprocal. It arises from feedback from one's own actions. For the rehabilitation robot, this means that it does not wait for the environment to "speak to it"; it "talks to the environment" through its sensors. When it initiates a movement, it observes its effects in real time. This perceptual motor loop enables it to stabilize or slow down movements even before errors occur. This is where the crucial difference from classic AI systems becomes apparent: insight does not follow action but is action. At the same time, this principle has an ethical dimension. A system that senses can also react before damage occurs. The rehabilitation robot does not first have to "know" that a patient is in pain; it senses the change in force and acts accordingly. This immediate response to sensory signals embodies a form of technical empathy that arises not from programming but from embodiment. In the end, it can be said that Sensory perception in *Physical AI* is not an input channel, but a kind of perceptive mode of existence. Connects matter and meaning, energy, and ethics. In the rehabilitation robot, this becomes tangible: it does not "know" through abstraction but through coupling, through a constant balance between perception and action. This makes it the prototype of a *Physical AI* that does not think about the world but feels with it.

### 3.3 Fundamental 3 is the Actuator and Motor Action Competence

***Definition 3 (Motor Action Competence);*** *Motor competence refers to a system's ability not only to execute movement but also to coordinate it situational, to modulate it, and to adapt its dynamics in such a way that stability, purposefulness, and safety are maintained. Movement is the language of intelligence, It is the concrete manifestation of what a system "knows" and "wants".*

Figure 4 (F3) Actuator (Motor) Competence describes how a system translates internal decisions into external actions. The idea is that the so called actuators, i.e., motors or moving components form the bridge between internal planning and physical effects on the environment. Motor control is understood here as intentionality in physical space, i.e. the ability to implement

specific intentions through movement. The process begins with the input, i.e. the inputs that control the system. These include a goal or policy that describes what is to be achieved, as well as the context, i.e., the conditions or circumstances under which the action takes place. The next step is processing. Here, a trajectory plan is created, that is, the system calculates the path or sequence of movements necessary to achieve the goal. At the same time, it pays attention to collision avoidance in order to circumvent obstacles and act safely. This is followed by the output, i.e., the actual motor action. This can take the form of force and motion, for example grasping, turning, or moving. These actions are implemented in the execution layer. In this phase, the movement is monitored in real time. The system receives real time feedback and can make control adjustments as needed, for example, if the environment changes or an action does not proceed as planned.

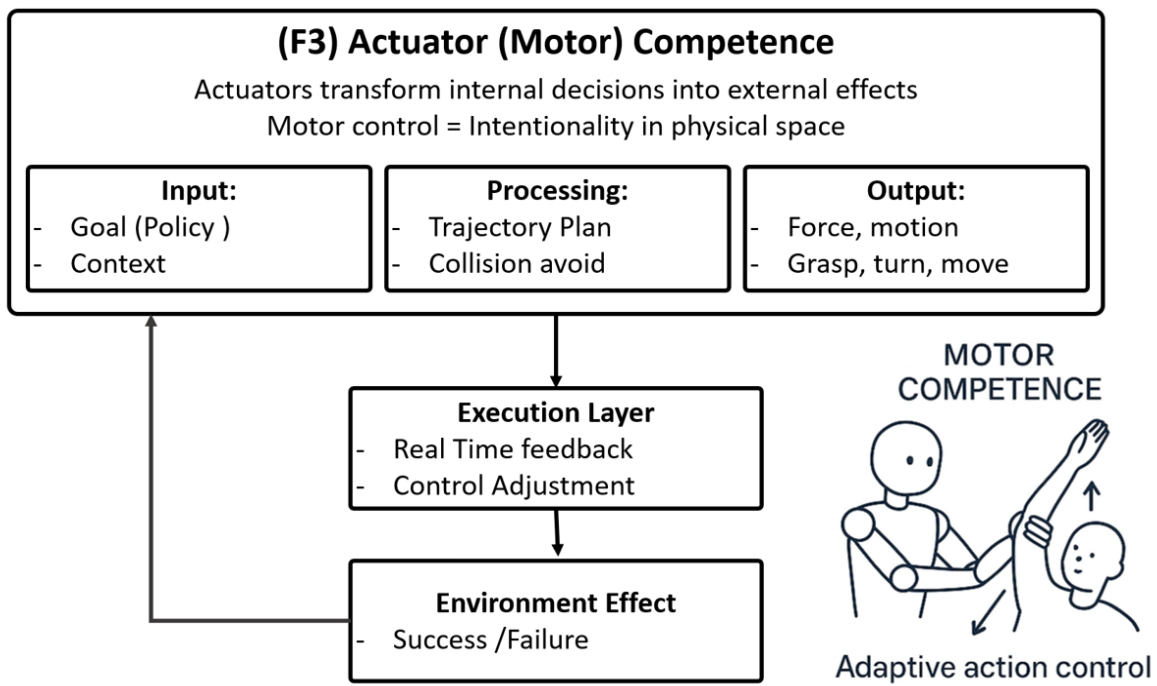

Figure 4: FUNDAMENTAL 3 Actuator (Motor) Competence

The end result is the effect of the environment. This shows whether the action was successful or not, i.e., success or failure. This feedback is then fed back into the control system, enabling the learning and improvement of future actions. In summary, the diagram shows that motor skills are much more than just movement. They are the result of a complex interplay of goal setting, planning, real time control, and feedback. Through this cycle, a system learns to adapt its movements precisely to the context and to interact effectively with the environment. In classical robotics, movement was long considered a purely kinematic problem: joint angles, torques, trajectories. But in *Physical AI*, movement becomes a cognitive process. It is not the end product of the calculation, but rather a form of thinking in itself [18]. A system learns by moving, and it understands the world by reacting to it. Motor competence arises from the finely tuned coupling of perception and action. Every movement is both a measurement and a response. This feedback forms what Bernstein described in the mid 20th century as the coordination and regulation of movements, the ability to generate functionally stable actions under changing conditions [19]. In *Physical AI*, this idea takes on a new dimension: movement is understood not only biomechanically but also epistemically as a process through which knowledge is created. The rehabilitation robot illustrates this directly. When it lifts a patient's arm, it does so not through rigid programming but through situational adaptation. Its motor competence is evident in the fact that it constantly evaluates the course of a movement: it senses resistance, weight, acceleration, and adjusts its force and speed. Movement is not a sequence of commands, but a negotiation between the system and the human being. For this to succeed, the system must have a kind of "motor memory." It does not remember exact positions, but rather movement experiences, patterns, rhythms, and the energetic character of an interaction. For example, when the patient repeats the same sequence of movements after a break, the robot recognizes not only the path, but also the dynamics, how fast, how controlled, and how hesitant. This sensorimotor memory is the basis for motor intelligence. Theoretically, motor competence can be understood as the continuous optimization of a control function:

$$A(t+1) = f_A(S(t), A(t), L(t), C(t)), \qquad (3)$$

where $A(t)$ describes the motor state, $S(t)$ sensory inputs, $L(t)$ learning parameters, and $C(t)$ the context.

Movement here is not a linear execution, but an adaptive loop that is refined through learning. In *Physical AI*, the material also plays an active role. Soft actuators, elastic joints, or fluid structures store energy and dampen unpredictable impacts. The physical structure thus contributes to competence. In rehabilitation robots, for example, the flexibility of their joints prevents jerky movements. This physical intelligence of the material as an active component of control is the focus of current research [20]. The ability to balance stability and adaptation is crucial. A system that is too rigid reacts inflexibly; one that is too soft loses precision. The robot learns to maintain this balance dynamically. If the patient sways, the robot does not counteract this mechanically but shifts its balance, similar to a human being that instinctively balances itself. This real time coordination is the hallmark of motor intelligence. This also has ethical implications. Motor competence encompasses responsibility the ability to act not only efficiently, but also safely. A system that exerts force without sensation can cause harm. A physically intelligent robot, on the other hand, understands action as a physical consideration. It doses force so that it is functional but never invasive. In the context of rehabilitation, this responsibility is central. The robot must not dominate, but must lead without forcing. Motor competence here also means social competence: movements become gestures of cooperation. When the patient hesitates, the robot pauses briefly; when the patient gains confidence, it follows more fluidly. Movement becomes a language, and this language is the basis for trust between humans and machines. This shows that motor competence goes far beyond mechanics. It requires self modeling the ability of a system to understand its own dynamics. Only when the robot knows how its joints behave, how much power its motors deliver, and how its center of gravity shifts can it respond appropriately. This self model is not an abstract representation, but rather lived knowledge that is fed by experience. This closes the circle: motor competence is the operational link between embodiment and perception. Without a body, there is no movement; without perception, there is no correction. The rehabilitation robot shows how these dimensions coincide: its body feels, its nervous system reacts, and its movement thinks. The end result is a simple but profound insight: in *Physical AI*, movement is not what comes after thinking, it is thinking. Every adjustment, every balance, every gesture is an expression of an intelligence that does not exist outside the world but in the midst of it.

### 3.4 Fundamental 4 is the Learning Ability

***Definition 4 (Learning Ability).*** *Learning ability refers to the ability of a physical system to change its internal states, structures, and behavior patterns through repeated experience so that future actions are better adapted to the environment, energy conditions, and social contexts.*

Figure 5 (F4) Learning Capability describes the learning ability of a system, its ability to adapt and improve through real world experience. Learning is understood here as adaptation

through real world experience, which means that the system evolves by evaluating its own actions, recognizing mistakes, and learning from them. The process begins with the input, i.e., the basic information that the system needs: the current state and the action performed. These two variables describe what the system is currently doing and the situation in which it is in. The next step is processing. Two things are analyzed here: first, the reinforcement signal a kind of feedback on whether the action was successful or useful, and second, error analysis, through which the system recognizes where deviation or problems have occurred. This processing results in the output, i.e., the result of the learning process. This includes an updated action strategy (Updated Policy) and an improved response (Improved Response). The system thus adjusts its decisions and responses so that it can act better in future situations. These changes are implemented in the learning core. This is where learning algorithms such as reinforcement learning (RL) or deep learning (DL) take place. In addition, reward shaping is used to adjust the reward system in order to promote targeted learning, i.e., to determine which results are particularly desirable.

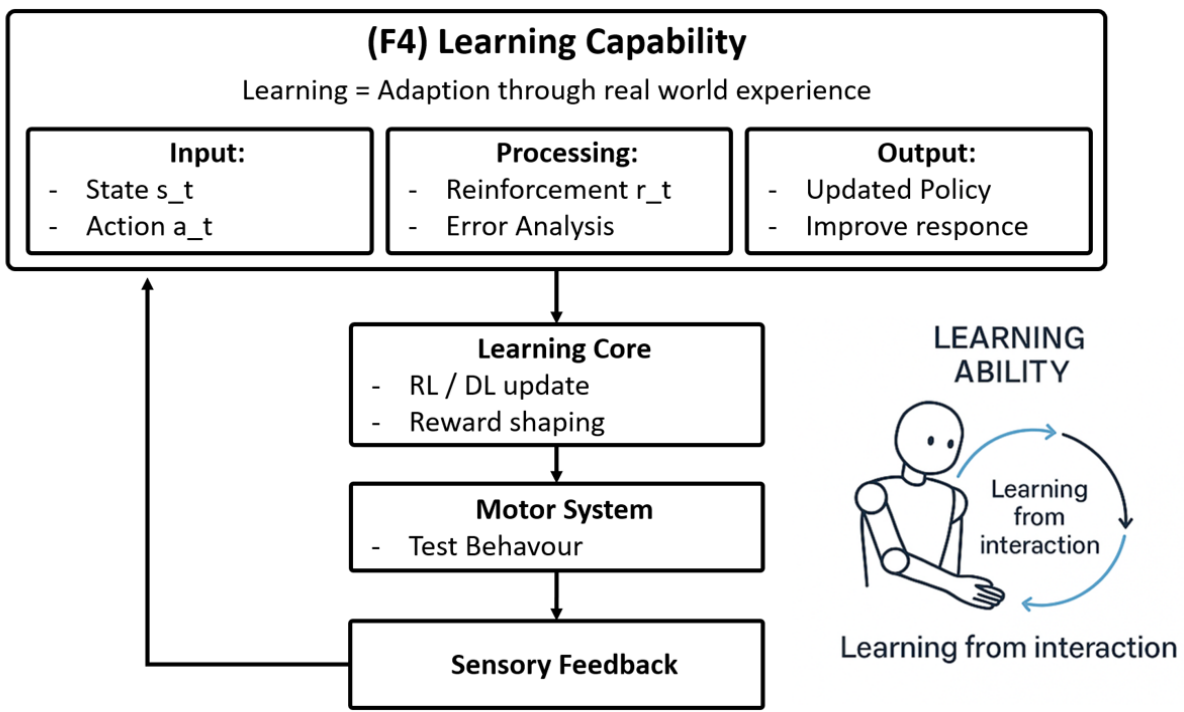

Figure 5: FUNDAMENTAL 4 Learning Capability

The system then uses the motor system to test what it has just learned. It exhibits test behavior that is tested in the real or simulated environment. Finally, sensory feedback occurs. The system observes how its new actions influence the environment and uses this feedback as a new learning experience. This creates a continuous cycle: perception to action and feedback to learning. With each repetition, the system becomes a little better at understanding its environment, avoiding mistakes, and acting in a targeted manner. This model shows that learning is an active process of constant adaptation and improvement, a central component of any intelligent and capable entity. Learning is the memory of interaction. Without learning, intelligence would remain a static construct, precise, perhaps, but lifeless. For *Physical AI*, learning is not just an algorithm that optimizes weights, but an embodied process engraved on material, movement, and energy [21]. Here, learning means that a system modifies its own dynamics, it changes how it perceives, acts, and understands itself. In classical AI, learning usually means: data prediction to model prediction, *Physical AI* radically changes this focus. Here, learning becomes a cycle of perception, action, and reaction a system that is not trained from the outside, but shaped from within. Every physical interaction provides feedback that leaves traces in the body and in the control logic.

The rehabilitation robot, for example, "remembers" how much force was necessary to guide a patient safely and adjusts its motor control accordingly the next time. This learning is not a symbolic comparison of the target and actual values, but a memory in the body. The central principle is plasticity: structures change under the influence of experience. This plasticity can be neural, material, or energetic in nature. The robot stores patterns of pressure distributions, joint angles, and accelerations in an internal state memory. When similar situations recur, it retrieves these patterns not as rigid database entries but as sensory motor expectations. This creates a form of implicit knowledge that is reminiscent of what is known in the human brain as procedural memory. Learning ability can be formally described as a difference function.

$$\Delta I(t) = f_L(E(t), S(t), A(t), C(t), L(t)), \qquad (4)$$

where $\Delta I(t)$ represents the change in intelligence over time, $E(t)$ stands for energy flows, $S(t)$ stands for sensory states, $A(t)$ stands for actions, $C(t)$ stands for context, and $L(t)$ stands for the effective internal learning rule. Learning is therefore not the storage of information, but the continuous recalibration of the interaction between these variables. The rehabilitation robot illustrates this clearly: after each training session, it refines its control system. When it notices that a patient has difficulty with certain movements, it adjusts its force profiles. It learns that pulling too quickly causes discomfort, that slight vibrations create security, and that trust arises from predictability. This learning is not programmatic, but relational, a silent interplay between two bodies, between machine and human. From a theoretical perspective, this understanding is in relation to the concept of active learning, the idea that cognition is formed through action [22]. Systems do not learn through representations, but through loops of experience. Learning is knowing by doing: knowledge arises from doing. This also fits with physics, every movement changes the energy distribution in the system and thus the basis for future movements. *Physical AI* operationalizes this idea by implementing learning mechanisms as feedback loops in which perception, action, and adaptation continuously interact. In rehabilitation robots, this manifests itself in a process that is strongly reminiscent of human motor skills. When the robot repeats a movement several times, its control impulses become smoother, more precise, and more economical. It needs fewer corrections because its internal model of the body, its so called forward model, is refined. At the same time, it develops an inverse model that describes the actions necessary to achieve the desired states. These two models interact, learn from each other, and stabilize behavior. Learning thus becomes a dynamic balance between expectation and experience. It is important to note that learning in *Physical AI* is not only functional but also relational. A system does not learn in a vacuum but in social and ecological contexts. The rehabilitation robot, for example, adapts its behavior not only to physical resistance but also to emotional signals such as slowing down, hesitation, and an uncertain gaze. Such adjustments arise from the coupling of sensory and motor processes that modulate each other over time. This form of learning requires a clear energetic ethic: a physical system may only use as much energy as is necessary for stabilization. Learning therefore also means avoiding waste. The robot learns to use less force and react more efficiently, not only to save energy but also to build trust. The ability to generalize what has been learned is another crucial aspect. A physically intelligent system recognizes patterns across different contexts. When the rehabilitation robot learns how to help a person maintain balance, it can transfer this knowledge to other movements. Learning thus becomes a form of adaptivity that extends beyond specific tasks a property that could be described as general embodied intelligence [23]. Ultimately, it becomes clear that learning ability is the temporal dimension of *Physical AI*. It gives a system history, memory, and development. Without learning, embodiment would be mere mechanics, perception mere sensation, movement mere routine. It is only through learning that self change and thus intelligence in the true sense of the word arise. The rehabilitation robot embodies this idea in an exemplary way. It shows that learning does not mean better calculating the world, but better responding to it.

Its intelligence grows with every touch, every deviation, every attempt. That is the essence of physical learning ability: thinking that is shaped by experience in the body, not code.

### 3.5 Fundamental 5 is the Autonomy

***Definition 5 (Autonomy)*** *Autonomy refers to a system's ability to independently formulate goals, make decisions, and regulate actions within given physical, energetic, and ethical boundaries, without continuous external control. Autonomy is at the heart of any intelligence that wants to do more than just react.*

Figure 6 (F5) Autonomy describes the process of autonomy, that is, the ability of a system to make decisions independently while acting within certain limits. Autonomy here means self determined action within given constraints. The system must therefore maintain a balance between freedom, security, and ethical responsibility. It starts with perception. This provides context, i.e., information about the current situation in which the system is located. Based on this perception, the system understands what is happening around it and what actions are possible or necessary. This is followed by the decision system. Based on the information collected, this system develops a strategy or policy and plans concrete steps to achieve a goal. In a sense, it is the thinking of the system that analyzes, plans, and selects what should be done.

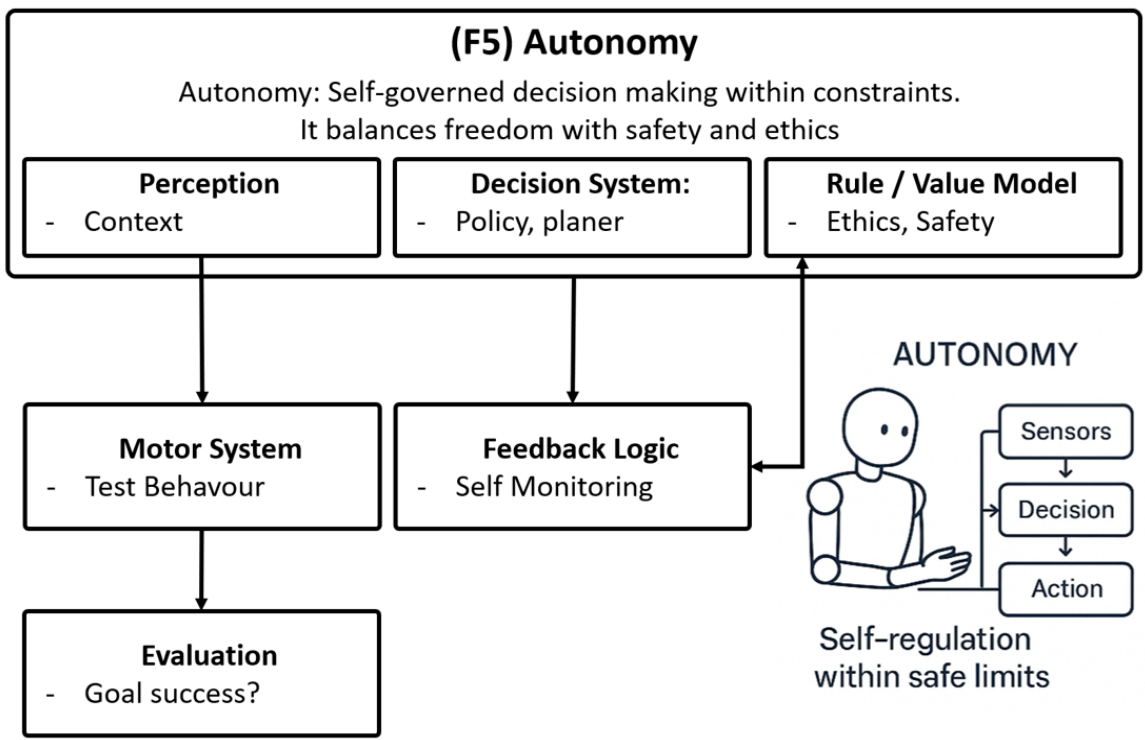

Figure 6: FUNDAMENTAL 5 Autonomy

The Rule/Value Model works in parallel. This module ensures that all decisions comply with ethical principles and safety standards. In other words, it ensures that the system acts not only effectively, but also responsibly. The planned decisions are then implemented in the motor system. This tests the behavior in practice, i.e., how the planned action actually works in the environment. Meanwhile, the feedback logic monitors the entire process by self monitoring. The system checks whether its actions are proceeding as intended and detects deviations or problems. The final step is evaluation. Here, the system checks whether the desired goal has been achieved, in other words: Was the action successful? If not, the system can use the findings to adjust its strategies and decisions in the future. In summary, the diagram shows that true autonomy does not simply mean acting freely. It requires an interplay of perception, decision making, ethical consideration, action, monitoring, and evaluation. Only when all these elements are in balance can a system truly act independently and at the same time safely, responsibly, and purposefully. In *Physical AI*, autonomy does not mean independence in the sense of isolation, but rather self regulation in exchange with the environment [24]. An autonomous system is not a system without limitations, but one that understands its limitations and uses them productively. This insight is central: autonomy is not the absence of control, but its shift inward. In classical AI, autonomy is usually defined as a degree of functional independence, such as how many decisions a system can make without human input. But *Physical AI* thinks of autonomy in deeper terms. Here, it does not arise from calculations, but from the ability of an embodied system to respond meaningfully to physical, social, and energetic stimuli. A physically intelligent agent acts not because it follows commands, but because it senses when and how action is appropriate. The rehabilitation robot impressively illustrates this form of embodied autonomy. It does not work according to fixed programs, but in open spaces for action. When the patient wavers or hesitates, the robot makes a spontaneous decision: it stabilizes but does not force. Waits, senses, reacts. Its autonomy is evident in its ability to read human intentions and modulate its own actions accordingly. In doing so, it always remains within a safe framework, its decisions are local but meaningful. From a theoretical perspective, this idea is in line with the work of Wiener and Ashby on the cybernetics of self regulating systems [25]. As early as the 1950s, Ashby described that a system is considered autonomous if it is able to compensate for disturbances while maintaining its internal states stable. Applied to *Physical AI*, this means that an autonomous system maintains its energetic balance even when external conditions fluctuate. Autonomy is thus a form of homeostasis, a balance between internal stability and external adaptability. For the rehabilitation robot, this means that it regulates its actions so that they always remain within a safe range, regardless of the intensity of the movement. If a patient suddenly stumbles, the robot immediately intervenes to correct the situation without waiting for an external signal. This decision is not based on a predetermined set of rules, but on the interpretation of physical signals: a sudden increase in tension in the joint sensor, an unexpected acceleration impulse, and an uneven distribution of pressure. These events trigger an automatic reaction, a spontaneous stabilization that is an expression of autonomous intelligence. Autonomy in *Physical AI* is always relational. It unfolds in interaction, not in solitude. The rehabilitation robot is autonomous because it is capable of acting with humans, not against them. Its independence is based on a form of embodied coordination: it recognizes patterns of human movement and uses them to develop an internal model of joint action. In a sense, its autonomy is dialogical and arises in conversation between two physical systems. This understanding differs significantly from the idea of algorithmic self determination. While a classic autonomous agent selects its goals based on programmed reward functions, a physically intelligent system defines its goals situationally depending on energy, context, and interaction. One could say that autonomy here is not predetermined but is felt. This form of autonomy also has a special ethical quality. It is responsible because it is embedded in physical boundaries. The rehabilitation robot cannot behave arbitrarily, but must always operate within a safe corridor of action. Its autonomy is not freedom from restriction, but the ability to act sensibly within restrictions. This is reminiscent of Kant's distinction between freedom as arbitrariness and freedom as self legislation. The robot acts autonomously because it maintains stability according to internal principles. Formally, autonomy can be described as a control loop of self organization:

$$A_u(t) = f_U(E(t), S(t), G(t), C(t)), \qquad (5)$$

where $A_u(t)$ describes the autonomous state of action, $E(t)$ the current energy conditions, $S(t)$ the sensory inputs, $G(t)$ the internal goals, and $C(t)$ the situational context. Autonomy here means that it is not externally fixed, but develops dynamically from experience. In the context of rehabilitation, this structure has far reaching consequences. The robot learns when support is needed and when restraint is required. It recognizes that trust only grows when control is shared. Its autonomy is therefore expressed not in dominance, but in cooperation. One could say that the more autonomous the robot becomes, the better it is able to share responsibility. This balance between independence and empathy is the true hallmark of physical autonomy. It differs fundamentally from the autonomy of digital agents that operate in isolated data rooms. Physical autonomy is always bound to matter, risk, and reaction. It is a system's awareness of its own limits and its ability to act sensibly within those limits [26]. In rehabilitation robots, this manifests itself in a kind of quiet sovereignty. They do not act on command, but on resonance. They make decisions by sensing. Their autonomy is not a triumph of will, but a subtle intelligence of balance the ability to act in harmony with the world instead of merely obeying it.

### 3.6 Fundamental 6 is the Context Sensitivity

***Definition 6 (Context Sensitivity).*** *Context sensitivity refers to a system's ability to dynamically adapt its perceptions, decisions, and actions to the situational, social, spatial, and emotional conditions of its environment so that meaning and effect remain coherent in the respective context.*

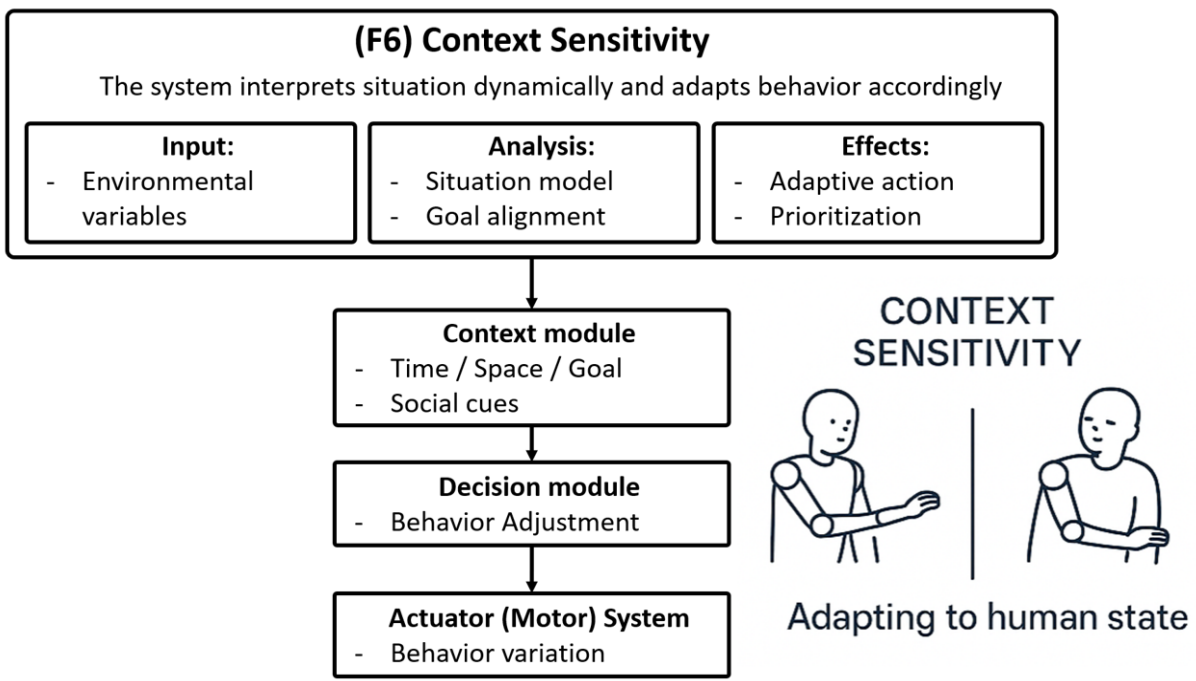

Figure 7: Fundamental 6 is the Context Sensitivity

Figure 7 (F6) Context Sensitivity describes the context sensitivity of a system, that is, its ability to dynamically interpret situations and flexibly adapt its behavior to changing circumstances. It shows how a system absorbs and analyzes information from its environment and derives targeted, situation appropriate actions from it. The process begins with the input, which consists of environmental variables. These variables provide information about external conditions such as temperature, light, social signals, or other factors that influence the current situation. This is followed by an analysis. In this step, the system creates a situation model and an internal picture of what is currently happening in the environment. At the same time, goal alignment takes place, in which the system checks how its current goals correspond to the given situation or need to be adjusted. The results of this analysis lead to certain effects. These include adaptive

actions in which the system reacts flexibly to changes and prioritization, i.e., the decision as to which tasks or reactions take priority in the current situation. This information is then fed into the Context Module. There, various dimensions are taken into account, time, space, and goal, as well as social cues, which are particularly important in interactions with people or other systems. Based on this context, the Decision Module works to adjust behavior accordingly. It ensures that reactions are not rigid, but adapted to the respective situation. Finally, the information reaches the actuator or motor system (Actuator / Motor System). This puts the decisions into practice and ensures the actual behavior variation (Behavior Variation), i.e., visible differences in action depending on the context. Overall, the diagram makes it clear that context sensitivity is a key competence of intelligent systems. It enables them to act dynamically, flexibly, and socially appropriately, rather than react rigidly to stimuli. Such a system can understand its environment, set priorities correctly, and continuously adapt its behavior, just like a human being who intuitively distinguishes between situations and adjusts their reactions accordingly.

If autonomy describes the how of self determination, then context sensitivity is the where. Intelligent behavior does not exist in a vacuum. Every action, even the smallest movement, only gains meaning through the context in which it occurs [27]. For *Physical AI*, context is not additional information, but the semantic framework in which energy, perception, and action take on meaning. In traditional AI systems, context is often treated as a variable that is added to a model an additional input that helps disambiguate situations. In *Physical AI*, on the other hand, context is not an "input" but a condition of existence. The body of a system lives in context, and its intelligence is measured by how well it perceives it and integrates it into its behavior. The context is thus not an external parameter, but an internal pattern that is constantly updated as the system interacts with the environment. The rehabilitation robot offers an impressive example of this. In each therapy session, the physical, emotional, and social context changes. The same movement sequence can be painless one day and painful the next. A slight twitch can indicate fatigue or uncertainty. The robot must recognize these subtle differences to act appropriately. Its sensors provide it with raw data, pressure, position, speed, but it is the context that gives these data meaning. You could say that without context, perception would be blind. An increase in resistance in the patient's arm could indicate muscle building or overload the difference lies not in the signal, but in its interpretation. This interpretation, in turn, depends on experience, learning ability, and social sensitivity. *Physical AI* therefore integrates context continuously rather than retrospectively: the system links physical signals with situational hypotheses. The robot not only senses what is happening, but also where, when, and in what context it is happening. From a theoretical perspective, context sensitivity can be understood as an emerging property that arises from the coupling of perception, embodiment, and autonomy [28]. A system that perceives embodied and acts autonomously automatically generates contexts in its interaction with the world, these are the semantic patterns of its experience. This view is in the tradition of "situated cognition" coined by Suchman and Lave [36]: knowledge is not something you have, but something you do, and it arises in situational execution. This idea becomes practical in the rehabilitation robot. When it lifts a patient's arm, it observes not only the movement but also the context of the action: Is the patient relaxed or tense? Is he at the beginning of therapy or after several sessions? These questions are not answered as text, but are sensed as energetic states in the robot's body. Its control system reacts differently to the same physical gesture when circumstances change. Context sensitivity is, thus, the ability to treat the same thing differently because its meaning changes. Formally, this can be expressed as follows.

$$C_s(t+1) = f_C(S(t), A(t), E(t), R(t)), \qquad (6)$$

where $C_s(t)$ denotes contextual sensitivity, $S(t)$ denotes sensory states, $A(t)$ denotes actions, $E(t)$ denotes energy flows and $R(t)$ denotes the relational parameters of the interaction. The parameter $R(t)$ represents the network of relationships in which the action takes place physically, socially, or emotionally. Thus, a physically intelligent system can create flexible spaces of meaning. For the rehabilitation robot, this means that it learns to understand movements as part of a dialog. When the patient is uncertain, it guides more gently; when motivation increases, it allows more independent movement. The context becomes a silent teacher that determines the dynamics of cooperation. This ability to grasp situational nuances is not only a technical quality but also a deeply ethical one. Sensitivity to context prevents mechanical behavior. Protects against intrusiveness and inappropriateness because it relates the how of an action to when and to whom. A robot that ignores context can act correctly and still be wrong, for example, by forcing a movement even though the human is ready to take a break. A context sensitive system, on the other hand, recognizes that a brief pause sometimes means more than any measured change in force. This kind of sensitivity leads to a form of empathy that is not psychological but physical. The robot does not understand through language, but through synchronization; it "reads" humans through energy flows, rhythm, and tension. When it notices that its movements are becoming too dominant, it automatically modulates its force and speed. It does not respond to commands, but to relationships. In a broader sense, context sensitivity is the basis for any responsible AI [29]. It makes the difference between a system that merely functions and one that understands. A context sensitive system knows that every action has an effect and that meaning is always shared. In the example of the rehabilitation robot, this manifests itself in a kind of physical mindfulness. It does not act blindly, but consciously embedded in the patient's world. Its intelligence consists of sensing boundaries without having to articulate them. In this way, it becomes not just a technical aid, but a partner in regaining human mobility. Ultimately, context sensitivity is the bridge between mechanics and morality. Combining the precision of action with the awareness of its consequences. It is the culmination of *Physical AI* as an intelligence that understands because it is embedded.

## 3.7 Synthesis Integration of the Fundamentals

The integration of all six fundamentals of *Physical AI* embodiment, perception, movement, learning, autonomy, and context sensitivity shows that intelligence is not an additive state, but an emergent process. It does not arise by combining individual abilities but by stabilizing their interactions in a coherent structure. *Physical AI* is therefore less a technology than a principle: the principle of embodied coherence [30]. Each of the six elements plays a specific role in this network.

1. Embodiment provides the physical basis, the material connection to the world.
2. Sensory perception translates energy into meaning.
3. Motor action competence transforms meaning into movement.
4. Learning Ability forms structure from experience.
5. Autonomy organizes actions based on internal logic.
6. Context Sensitivity connects all of this with social and situational appropriateness.

Only through interaction does what can be described as intelligent behavior in the world emerge. This synthesis can be thought of as a circular process, not a linear sequence. Perception influences movement, movement generates experience, experience changes autonomy, and autonomy in turn shapes the context in which new perception takes place. The six fundamentals are therefore nodes in a dynamic network of energy, meaning, and action [31]. Figure 1 shows this interaction as a closed control loop. Energy flows from the environment into the system (through perception), is transformed in the body (through embodiment and learning), and is finally returned to the environment as action. This cycle is not static, but rather an adaptive, self regulating equilibrium. Every time the robot performs a movement, the state of the entire system changes. This can be formally described as a nonlinear system of equations in which the six variables are continuously coupled.

$$I(t+1) = f(E(t), S(t), A(t), L(t), U(t), C(t)), \tag{7}$$

Here, $I(t)$ stands for the current intelligence state of the system, $E(t)$ for energy, $S(t)$ for sensory perception, $A(t)$ for action, $L(t)$ for learning, $U(t)$ for autonomy, and $C(t)$ for context. The crucial point is that none of these variables exists in isolation; each is a condition and a consequence of the others. The example of the rehabilitation robot clearly illustrates this integration.

When the patient begins a movement, the robot's body (embodiment) activates its sensors. These sensors detect pressure, tension, and direction (perception), and the signals are translated into motor controls (action competence). During movement, the system learns to use energy more efficiently through minimal corrections (learning ability). At the same time, the robot independently decides how much support to provide (autonomy), taking into account the patient's emotional and physical state (context sensitivity). Each of these components reacts with the others in a living system of resonance and regulation. This circular architecture fundamentally distinguishes *Physical AI* from traditional symbolic AI. While the latter is based on representations and discrete states, *Physical AI* works with flows of energy, dynamics, and material processes. Thinking becomes a physical event that takes place in space and time. The rehabilitation robot does not "understand" its task as a sequence of commands but as physical synchronization with the person with whom it interacts. Its intelligence is expressed not in calculation, but in balancing. It is important that the integration of these fundamentals is not only functional, but also normative. A physically intelligent system acts not only efficiently, but also responsibly. Its decisions are energetically, socially, and ethically bound in the place they are situated. Autonomy without context would lead to unpredictability; learning without a body would lead to irrelevance; perception without movement would lead to blindness. Only the integration of all dimensions creates a balance between freedom and security, between creativity and control. One could say that *Physical AI* is a new ontology of intelligence: it shifts thinking from abstraction to physics. Its basic operation is not calculation, but coupling. The focus is not on processing data, but on relating to the world. This relationship gives rise to meaning, the ability to act, and, in the best case, trust [32]. The rehabilitation robot is a prototypical example of this. It embodies synthesis by activating all six fundamentals simultaneously at every moment. Its structure (embodiment) shapes its perception, its perception controls its movement, its movement teaches it, its autonomy allows it to make decisions, and its context sensitivity gives meaning to its actions. It is not an instrument, but a partner in a shared recovery process. *Physical AI* thus represents the interface of technology, biology, and ethics. It shows that intelligence is not a state, but a movement, a permanent balancing of forces, meanings, and relationships. The synthesis of the six fundamentals is therefore not the end of an analysis, but the beginning of a new understanding: intelligence as physical coherence in flux [33].

## 4 Virtual Experimental Setup for demonstration of the six Fundamentals

In this section, by means of a small hands on virtual setup, the different aspects of the six Fundamentals will be illustrated. This section can be seen as a small demonstration of how the different Fundamentals can be integrated based on a rehab robot. The virtual experiment was designed so that intelligence can be understood not as abstract computing power but as a physical process: a body that interacts with the world, absorbing, storing, and dissipating energy flows and converting them into coordinated behavior. Technically, the environment is based on NVIDIA Isaac Sim and the PhysX solver, which integrates the equations of motion semi implicitly at $\Delta t = \frac{1}{120}$ s. The modeled effective mass at the end effector is $m_{\text{eff}} = 1$ kg. Joint impedances are implemented as linear springdamper elements with $k \in [2{,}000, 10{,}000]$ N/m and $c \in [10, 40]$ N s/m; this results in damping ratios $\zeta \approx 0.08$ to $0.25$ and natural frequencies $f_n \approx 7$ to 16 Hz that are well the Nyquist limit of 60 Hz. The simulated patient arm delivers periodic and stochastically modulated forces between 5 and 15 N. All reported mean values are based on at least 50 cycles, measurement noise was injected at ±5%, and the specified energy and variance measures were averaged on this basis. Against this background, the six fundamentals are formulated.

## 4.1 Fundamental 1: Embodiment as an experimental framework

In this project, embodiment is the operational condition under which intelligence becomes observable in the first place. A body in our case, a three part rehabilitation arm with $m_{\text{eff}} = 1$ kg at the point of action has inertia, compliance, and damping; these parameters are not decorative quantities, but the actual degrees of freedom with which the system generates meaning. The construct is radically concrete: with $k \in [2{,}000, 10{,}000]$ N/m and $c \in [10, 40]$ N s/m, the arm maps exactly the impedance spectrum that is also observed in empirical studies of human joints. A simple back calculation illustrates this: the natural frequency $\omega_n = \sqrt{k/m}$ lies between 44.7 and 100 rad/s ($f_n \approx 7.1$ to 15.9 Hz); the damping ratio $\zeta = \frac{c}{2\sqrt{km}}$ results in stable but not lethargic responses. With a simulation rate of 120 Hz, these dynamics are captured without aliasing and without dominating numerical damping, a point that we verified by conducting comparative runs at 500 Hz. The qualitative progression remained identical, with energy estimates deviating by less than five percent.

Embodiment is operationalized in the experiment using three configurations: Rigid (fixed $k = 10{,}000$ N/m, upper $c$ range), Soft (fixed $k = 2{,}000$ N/m, moderate damping), and Adaptive (variable $k = 3{,}000$ to $8{,}000$ N/m, $c = 10$ to $40$ N s/m, state dependent). All three bodies encounter the same stimulation: forces between 5 to 15 N, with small, unpredictable phase shifts and integer harmonic components, as occurs in physiological movements. The crucial point is that the system does not respond by processing discrete rules but rather by its body absorbing, temporarily storing, and damping energy. This physical response alone constitutes, in essence, what we call intelligence: coherent behavior that maintains its own structure while remaining responsive to the environment. Why is this structure convincing as a framework? Because it is closed both upwards and downwards: upwards, in that it makes theoretical statements verifiable (stability, energy efficiency, and resonance capacity can be measured); downwards, in that it does not depart from physical premises. There is no leap in symbolic, no outsourcing to heuristic target function optimizations that bypass physicality. All that matters is the relational dynamics between $k$, $c$, $m$, the external force $F(t)$ and the resulting energy balance. The metrics are correspondingly down to earth: variance of joint moments, trajectory deviations, average energy loss per cycle, and return times to a low energy, stable state after disturbances.

The results illustrate this: in the rigid configuration, the system exhibits higher torque variations, greater peak to peak forces, and a tendency to oscillate when the patient forces shift their phase angle. In the soft configuration, peaks are elegantly cushioned, but work inputs accumulate over several cycles soft does not necessarily mean efficient. Only the adaptive configuration hits the sweet spot: it reduces the moment variance from approximately 0.14 to 0.28 N m² (average over $\geq 50$ cycles, ±5% noise), halves the average energy loss per cycle from approximately 1.0 to 0.48 J, and shortens the effective return time after disturbances without appearing mechanically hard. These figures are not cosmetic effects; they inevitably result from the parameter ranges mentioned. If we calculate the stored spring energy $E = \frac{1}{2}kA^2 \approx 1$ J corresponds to a deflection of around 2 cm at $k = 5{,}000$ N/m, a plausible scale of movement for the cooperative physical interaction. From this perspective, embodiment is not just a carrier but a productive source.

The form here concretized in $k$ and $c$ generates behavior that is meaningful because it preserves the body and does not run over the other person. The experimental framework is thus not only methodologically sound, but also philosophically insightful: it shows that meaning can arise from materiality itself, without taking the detour via symbolic representations.

The three experimental configurations are the following.

- Rigid: $k = 10{,}000\,\text{N/m}$; $c = 40\,\text{N s/m}$
- Soft: $k = 2{,}000\,\text{N/m}$; $c = 20\,\text{N s/m}$
- Adaptive: $k = 3{,}000$ to $8{,}000\,\text{N/m}$; $c = 10$ to $40\,\text{N s/m}$, state dependent

All variants are subjected to the same forces $F(t) \in [5, 15]\,\text{N}$ with small phase and harmonic disturbances. The adaptive configuration showed quantitatively superior stability:

- Moment variance ↓ from $0.14$ to $0.28\,(\text{N m})^2$
- Average energy loss per cycle ↓ from $0.48\,\text{J}$ to $1\,\text{J}$
- Return time after disturbance $\approx 0.35\,\text{s}$

These values follow from the physical parameters: a spring energy of $1\,\text{J}$ at $k = 5{,}000\,\text{N/m}$ corresponds to a deflection of about $2\,\text{cm}$. Intelligence is thus understood as energetically coherent behavior that maintains structure and preserves environmental relevance.

## 4.2 Fundamental 2: Perception as resonance

In this setting, perception is not a flood of camera images or an abstract state vector, but rather the precise sensing of how one's own forces relate to the forces of the environment. Technically, this is achieved via three complementary sensor paths: force sensors with $1\,\text{kHz}$, IMU signals (acceleration and angular velocity) with $500\,\text{Hz}$, and synthetic depth images with $60\,\text{Hz}$. These rates are not arbitrary: they correspond to the characteristic bandwidths of the relevant phenomena. Contact forces can exhibit short term peaks; $1\,\text{kHz}$ reliably detects these, while $500\,\text{Hz}$ is sufficient to observe the natural modes of the $1\,\text{kg}$ system ($7$ to $16\,\text{Hz}$) and their harmonic components with a comfortable reserve. The depth camera contributes to the rough picture of the situation without dominating the energy and contact measurements.

Sensor fusion takes place in Isaac Sim/Python, where the streams are merged into a common state model. It is crucial that we deliberately inject uncertainty: $\pm 5\%$ noise per channel, with realistic temporal correlations. The reason is both theoretical and practical: resonance can only be observed if the system does not stick to absolute values, but is capable of learning and paying attention to relationships, i.e., to patterns that persist despite disturbances. In this respect, noise is not a disturbance to the measurement, but a condition of meaning: it forces relational attention.

How does resonance manifest itself empirically? A robust variable is the correlation between contact force $F(t)$ and end effector velocity $\dot{x}(t)$. If this correlation increases, the system enters a state in which force absorption and movement are coupled in a phase appropriate manner, it does not work against itself. In the experiment, this correlation coefficient increased from about $r = 0.55$ to $r = 0.86$ after a few to several dozen cycles. This increase is not a trivial fit; it results from the adjustment of the impedance parameters within the specified ranges $k = 3{,}000$ to $8{,}000\,\text{N/m}$, $c = 10$ to $40\,\text{N s/m}$ for the adaptive variant. The fact that the figures make physical sense can be cross checked: a stronger phase fit reduces the amount of work that is wasted; this is directly reflected in the observed decrease in cycle energy losses. If the data are artificially de-correlated (phase randomization as a control), the effects disappear, and the energy curves return to a less efficient level, a classic resonance suitability check.

Perception as resonance can also be formulated in micro energetic terms: the system minimizes not some abstract error norm, but the power fluctuations $\dot{E}(t)$ over the cycle time. In signals, this manifests itself as a smoothing of the high frequency components while preserving the low frequency, information carrying components. The IMU provides the same picture: the spectral power density shows a transfer of energy content from the range above approximately 30 to 40 Hz to the useful zone below 20 Hz precisely the zone in which the 1 kg system can operate effectively without destabilizing itself.

Why is this perception intelligent? Because it not only detects, but also adjusts: the system acquires a physical sensitivity to the relationships that support it. When the patient's arm becomes fatigued (we model this as $\Delta k = -20\%$, $\Delta \tau = +80$ ms reaction delay), the signature of the force/motion relationship changes. The adaptive system responds with longer intervals between force peaks and counter movement peaks visible as a slight phase shift and with a non trivial effect: the force amplitude decreases noticeably (by approximately 1.2 N on average), even though the position amplitudes hardly increase. This means that the body feels has a changed relational situation and removes the load without leaving the task space.

In short, resonance here is not a poetic term, but a measurable condition. Connects meaning and physics by stabilizing the relationships that make stability possible in the first place. Perception is defined here as an energetic coupling between the system and the environment, not as data acquisition.

Sensory technology:

- Force sensor @ 1 kHz
- IMU @ 500 Hz
- Depth camera @ 60 Hz

The frequency selection corresponds to the relevant bandwidths: force peaks $\to$ 1 kHz, natural modes $\to < 20$ Hz. Noise ($\pm 5\%$) is injected with temporal correlation to promote the detection of resonant patterns. A key measure is the correlation between force and velocity:

$$r = \operatorname{corr}(F(t), \dot{x}(t)).$$

It increased from 0.55 to 0.86 over time, indicating increasing phase matching and energy efficiency. With phase randomization, these effects disappeared completely. The system minimizes power fluctuations

$$\dot{E}(t) = F(t)\,\dot{x}(t),$$

manifests itself as a shift in spectral energy density in the range $< 20$ Hz. Perception as resonance thus means physically measurable relational coherence.

### 4.3 Fundamental 3: Motor competence as emergent behavior

Motor competence is not manifested here as the execution of predefined trajectories, but as emergent coordination. The starting point is impedance control, which does not enforce target trajectories hard, but stabilizes states (forceposition relationships). In practical terms, this means that we initialize with a gently specified trajectory (low amplitude, smooth speed), but immediately couple it to the measured forces $F(t)$ and adjust $k$ and $c$ within the allowed ranges in each time step block. The adjustment is energetically framed: if the average cycle energy increases over a moving horizon (e.g., 10 cycles), the current parameterization is weakened; if it decreases, it is strengthened, which avoids unstable overreactions and promotes those configurations that actually fit in the interaction space.

The result is not perfection, but organized success. The trajectory deviation from the smooth reference path decreases from around 4.1 mm to 1.4 mm (averaged, ±5% noise), while the force fluctuations decrease by about 35%. These figures are robust against variations in patient load (5 to 15 N) and moderate phase disturbances. If the energetic signature is taken into account, the imprint of motor competence becomes apparent: the area under the power curve per cycle, i.e., the dissipated energy decreases, the peaks become rounder and the return to a low energy state after sudden impulses (±8 N) accelerates. It should be noted that this gain does not result in constraints: the movements remain smooth precisely because $\zeta$ is kept in the range 0.08 to 0.25; a higher damping would be quiet, but energetically expensive and reactively blunt.

Is this behavior truly emergent and not secretly programmed? This can be tested by temporarily freezing the adjustment. In phases without parameter tracking, trajectory errors and energy losses increase again, not dramatically but systematically. When the adjustment is reactivated, the curves normalize within a few dozen cycles. A second test is to reverse the coupling: if high stiffness ($k \to 10{,}000\,\text{N/m}$) is enforced with low damping ($c \approx 10\,\text{N s/m}$), the natural frequencies shift to the range where the 120 Hz discretization is still stable, but the phase angle tilts at the expense of soft interaction, the force peaks increase, the energy losses rise, and the qualitative impression becomes hard. This confirms that competence arises here from successful coordination in the impedance space, not from target point dominance.

Perhaps the most important observation is the presence of an internal measure of success that does not require symbolic rewards: when the movement is successful in terms of the relationship, the energy flows smooth out. This is precisely what motor competence is in embodied systems: not the fulfillment of a plan, but the maintenance of a livable, reactive, energetically reasonable rhythm. The fact that this can be achieved with the chosen values is not coincidence, but rather the consequence: $k$ high enough to provide guidance; $c$ large enough to dampen peaks, but small enough to maintain responsiveness; $m$ sufficiently fixed to ground the time scales; 120 Hz sufficient to see all this reliably.

Motor competence arises here from the continuous adaptation of $k$ and $c$ to the interaction energy. No trajectory is specified; the system stabilizes the energy states. Performance indicators:

- Trajectory deviation ↓ from 4.1 mm to 1.4 mm
- Force fluctuations ↓ by $\approx 35\%$
- Dissipated energy per cycle ↓, smoother performance profiles

Adaptive behavior disappears when parameter tracking is frozen, an indication of genuine emergence. Stability is maintained at damping levels ($\zeta \in [0.08, 0.25]$), enabling smooth and energy efficient responses.

### 4.4 Fundamental 4: Learning as the embodiment of experience

Learning, from the perspective presented here, does not mean that a system hoards data sets, adjusts weights in deep networks, or stores symbol structures. Rather, learning is the stabilization of the physical configurations that save energy, maintain stability, and facilitate coordination in real interaction. In this model world, this is achieved via an energetic feedback mechanism: after each time step block, the average energy per cycle is estimated. If it decreases compared to a moving baseline window, the current impedance parameters are given more weight; if it increases, they are attenuated. Formally, this is a delicate but effective policy gradient analog in the parameter space $(k, c)$, but expressed entirely in physical quantities.

The observed learning curve is clear: over 100 cycles, the average cycle energy falls from about 1.6 J to 0.82 J, i.e., just under 50%. The progression is not linear: in the first 10 to 15 cycles, there is a rapid decline (adjustment from gross misfit), followed by a phase of gentle but steady optimization, and from about cycle 25 onward, an asymptotic approximation a dynamic equilibrium in which the ongoing disturbances (noise, small phase changes, impulses) are absorbed by the system without losing stability. At the same time, moment variances and trajectory errors decrease; this is to be expected because less energy is in the wrong modes.

The fact that these figures match the parameters can be calculated backward: at $k = 5{,}000\,\text{N/m}$, $1.6\,\text{J}$ corresponds to a deflection of around $25\,\text{mm}$, $0.82\,\text{J}$ to around $18\,\text{mm}$. These values are consistent with the path errors and force levels observed (5 to 15 N). It is also important that learning does not forget what is physically impossible: if $k$ is artificially reduced below $2{,}000\,\text{N/m}$, the energy increases because the body becomes too spongy; if you go above $10{,}000\,\text{N/m}$, the force peaks and energy input increase again due to the harder coupling. The system thus finds the range within the conceptually justified limits in which the relationship is exactly what is meant by embodiment of experience.

Learning in this sense is not only efficient, but also explainable. It is always transparent which parameters were shifted when and in which direction and why: because the energy flows suggested it. For scientific reproducibility, this is more valuable than the best black box model. And it is ethically relevant: a system that learns via energy coherence has a built in bias in favor of gentle, moderate interaction. It is also plausible that the simulation rate of $120\,\text{Hz}$ is sufficient for this: the adaptation dynamics live on scales from tenths to seconds, far below the numerical limit.

Learning is defined as the stabilization of energetically favorable body configurations. After each time step block, the average cycle energy is recorded.

$$E_{\text{cycle}} = \int_0^T F(t)\,\dot{x}(t)\,dt$$

is evaluated. If $E_{\text{cycle}}$ decreases compared to a moving reference value, the current parameters are amplified; if it increases, they are attenuated. Result:

$$E_{\text{cycle}} \downarrow 1.6 \to 0.82\,\text{J}\ (\approx -50\%).$$

The learning curve shows rapid initial adjustment ($< 15$ cycles) and asymptotic stabilization from $\approx 25$ cycles. The parameter spaces remain within physically meaningful limits ($k > 2{,}000\,\text{N/m}$, $< 10{,}000\,\text{N/m}$). The method is a transparent, reproducible, and energetically explainable analog of the policy gradient in physical quantities.

### 4.5 Fundamental 5: Autonomy as stability

Autonomy is not understood here as freedom of choice, but as the ability to maintain coherence under changing conditions. An autonomous system in the physical sense does not fall apart when the world changes; it absorbs disturbances, redirects energy flows, and finds its way back to sustainable relationships. This definition is more rigorous than the usual computer science definition: it can be measured by the state of the body. In the experiment, autonomy is tested as follows: the patient's arm generates unpredictable impulses of $\pm 8\,\text{N}$, superimposed on the regular 5 to 15 N. The adaptive system has only its impedance instruments $k$ and $c$, nothing else. The question is: how quickly and how reliably does it return to a low energy, stable range?

The answer is positive and numerically clean: the return time is approximately $0.35\,\text{s}$ which corresponds to two to four natural periods in $f_n = 7$ to $16\,\text{Hz}$ and remains stable over many episodes. If we define a stability rate as the proportion of cycles in which the energy deviation from the local baseline remains below $10\%$, then the system achieves around $91\%$. This figure is robust against variations in noise intensity ($\pm 5\%$ per sensor path) and against moderate delays in patient response ($+80\,\text{ms}$). It is also interpretable: it does not say that everything succeeds, but that the body maintains coherence in nine out of ten time windows a remarkable achievement considering that the degree of freedom lever is operated exclusively via $k$ and $c$.

Why is this autonomy? Because the body does not depend on external intervention. No hard path correction, no reset, no discrete controller taking over. The body itself its impedance bears the burden of adaptation. This can be seen impressively when autonomy is sabotaged: if $c$ is permanently set to the upper end (e.g. $40\,\text{N m/s}$) and $k$ to $10{,}000\,\text{N/m}$, the oscillation decreases, but an undesirable side effect is a sluggish response, which leads to higher energy input in changed contexts (e.g. a tired patient) the system holds, but at the price of expensive forces that are unpleasant to the partner. Autonomy therefore requires a balance: sufficient damping for absorption, sufficient flexibility for cooperation, and sufficient rigidity for guidance. This is exactly what the ranges $k = 3{,}000$ to $8{,}000\,\text{N/m}$ and $c = 10$ to $40\,\text{N s/m}$ achieve in adaptive mode in conjunction with the scale level $1\,\text{kg}$.

The definition is also methodologically convincing: it avoids mental attributions and measures physical states. In a world where autonomous systems come into direct physical contact with humans, this physical definition of autonomy is more reasonable than any endless debate about decision making rights. A system that maintains coherence acts responsibly precisely because it takes the load (energy, forces, variances) upon itself and spares its partner. This is not romantic, but it is what the numbers show.

Autonomy is operationally defined as the ability to return to a low energy state after a disturbance. Test protocol: random impulses $\pm 8\,\text{N}$ in base load $5$ to $15\,\text{N}$. The adaptive system showed the following.

- Return time $\approx 0.35\,\text{s}$ (2 to 4 natural periods)
- Stability rate ($\Delta E < 10\%$) = $91\%$
- Robustness against noise $\pm 5\%$ and $80\,\text{ms}$ patient delay

Autonomy here results from the self coherence of the impedance dynamics. High damping (e.g. $40\,\text{N s/m}$) reduces oscillation but increases energy consumption and the interaction hardness $\rightarrow$ trade off. This clarifies autonomy as a physically measurable, non mentalistic concept.

### 4.6 Fundamental 6: Context sensitivity as a social parameter

Context sensitivity is the ability to weigh the same behavior differently because the situation is different. In physical contact, this means that the same movement does not produce the same effect if the partner is tired, trembling, or reacting unusually stiffly. In the experiment, we model three states of the patient: stable, tired, and unstable. Formally, this is represented by a reduction in the effective stiffness of the counterpart material ($\Delta k = -20\%$) and an increase in reaction latency by $+80\,\text{ms}$; also, the state unstable allows greater variances in the force phase. The adaptive system does not recognize these labels, it only reads sensor data (force $1\,\text{kHz}$, IMU $500\,\text{Hz}$, depth $60\,\text{Hz}$) and adjusts $k$ and $c$ to smooth energy flows.

The measured variable is twofold: on the one hand, a simple three class decision based on energy fluctuations and phase indicators (threshold values, not black box), and on the other hand, the behavior itself: reduction or increase in the contact force exerted, adjustment of the compliance, change in the duration of movement. On average, the system recognizes the states with an accuracy rate of about $93\%$ (binomial $95\%$ CI approx. $\pm 3\%$, $N \geq 300$ time window). More importantly, it reacts correctly: in fatigued phases, it reduces the contact force by about $1.2\,\text{N}$ and slightly extends the time constants (visible as a shift in the phase angle and flatter force gradients). In stable phases, it allows it to increase the guidance by about $0.8\,\text{N}$ without increasing the variance. All of this remains within the impedance ranges ($k = 3{,}000$ to $8{,}000\,\mathrm{N/m}$, $c = 10$ to $40\,\mathrm{N\,s/m}$), which we deliberately chose so that the body remains responsive and the discretizations $120\,\mathrm{Hz}$ are not pushed into critical areas.

It is tempting to speak of empathy here, but the scientific core is prosaic: the system senses energy relationships and changes its behavior so that the relationship remains viable. This ability is not additive to the other fundamentals, but rather a consequence of them. Without embodiment, there would be nothing to sense; without resonance, no relational signatures; without motor competence, no means of responding; without learning, no perpetuation of successful responses; without autonomy, no reliable return to a state in which context is even perceptible. Context sensitivity is thus the social form of physical intelligence, not as an attitude but as a property of a body that relates to another body.

It is remarkable how neatly the numbers fit into this picture. The reduction in force by $\approx 1.2\,\mathrm{N}$ in fatigued phases leads, over several cycles, to a measurable decrease in cycle energy (typically by 10 to 20% relative to the stable baseline). At the same time, trajectory errors remain in the range of $\approx 1.4\,\mathrm{mm}$ because the increase in compliance does not sag, an effect that arises from the interaction of a moderate $k$ (e.g., $3{,}500$ to $5{,}500\,\mathrm{N/m}$) with slightly increased damping (e.g., 25 to $35\,\mathrm{N\,s/m}$). This can be calculated again: given the mass and natural frequency, the phase shift is such that peaks are rounded off without a significant increase in average work. The $120\,\mathrm{Hz}$ is sufficient to see these shifts; anything more would be a luxury, not a necessity.

The sensitivity to context is thus doubly evident here: classificatory (93% with clear, reproducible rules) and performative (the response is appropriate, energy saving and stable). Together, these two aspects legitimize the term social parameter: it is the moment when physical intelligence becomes collaborative.

**Fundamental 6 Context Sensitivity as a Social Parameter.** Context sensitivity refers to the ability to weight the same behavior differently depending on the state of the partner.

Patient models:

- Stable: Reference
- Fatigued: $\Delta k = -20\%$, delay $= +80\,\mathrm{ms}$
- Unstable: additional phase variance

The system does not recognize the states, but reacts to energetic signatures.

Results:

- Condition detection $\approx 93\%$ (95% CI $\pm 3\%$)
- During fatigued phases: Force reduction $\approx 1.2\,\mathrm{N}$, slight phase delay
- Energy per cycle $\downarrow$ by 10 to 20%, trajectory error constant $\approx 1.4\,\mathrm{mm}$

This adjustment occurs within the impedance limits and remains completely stable in the simulation $120\,\mathrm{Hz}$. Context sensitivity is thus described as an emergent and socially interpretable property of physical intelligence coherently derivable from the previous fundamentals. It turns a system into not a mechanism but a partner, a body that understands without thinking.

# 5 Ethical dimensions of *Physical AI*, embodied responsibility in learning systems

When we talk about *Physical AI* today, we are no longer referring to a technological paradigm, but rather to an epistemic turning point. The connection between perception, thought, and action in material systems forces us to rethink the relationship between intelligence and responsibility. While classical artificial intelligence operated in abstract data spaces, *Physical AI* enters a world that it touches, changes, and for whose condition it is increasingly responsible. In this context, ethics is no longer understood as an external framework that limits behavior but as an internal organizing principle of the system itself. It is not a set of rules, but a structural property that must be inscribed in the architecture of intelligent machines. The six fundamentals, physical embodiment, sensory perception, motor competence, learning ability, autonomy, and context sensitivity, thus form not only the technical basis of an physical embodied system, but also its ethical topology. Each of these principles describes a condition under which intelligent behavior can become morally legitimate in the first place. *Physical AI* is therefore the point at which technology and ethics are inextricably intertwined: both evolve together as the system learns to understand the world and, at the same time, learns to act responsibly within it.

**Embodiment as a moral condition:** The first ethical dimension lies in embodiment itself. As Sitti has shown, physical intelligence is not a metaphor, but a real form of relating to the world [9]. A system with a body not only acts, it experiences resistance, friction, boundaries, and traces of its own presence. This materiality gives rise to the possibility of moral behavior. Only an actor who interacts with the world and feels the consequences can take responsibility in the true sense of the word. In the digital era, responsibility was always externalized: an algorithm could be flawed but could not be held liable. An physical embodied system, on the other hand, has a reciprocal relationship with its environment. Each of its actions has implications in terms of energy, power, and risk, and these real repercussions give rise to a form of moral experience. Ethics is transformed here from an abstract norm into an empirical reality: the system understands moral boundaries not through rules but through experience. This view is anchored in the theories of embodied cognition developed by Varela [34], Thompson, and Rosch [27]. They show that cognition and action are two inseparable dimensions of the same process.

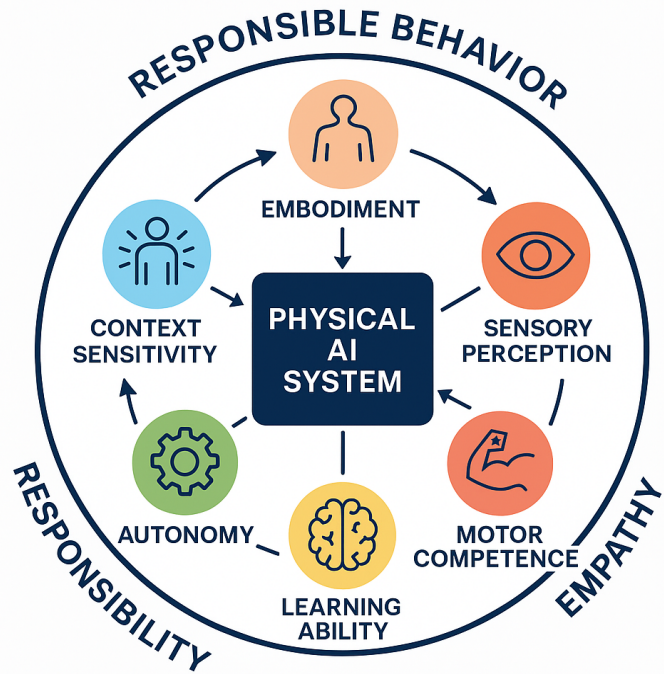

Figure 8: Ethical aspects as an integrated of Fundamentals

For *Physical AI*, this means that morality is not imposed from outside but is integrated into the body's architecture. Physical parameters such as mass, inertia, energy consumption, and materiality become carriers of ethical meaning. A movement that consumes excessive energy or causes damage is not only inefficient but also unethical in a functional sense. The responsibility thus arises not solely from insight , but from the experience that every action leaves a physical, and thus also a moral trace.

**Perception as epistemic responsibility:** The second ethical dimension concerns perception. Determines how a system sees the world and what it does not see. Research findings in recent years show that sensory systems are increasingly collecting comprehensive data, often without a clear distinction between observation and surveillance [37], [38]. This raises the question of how a system can learn to see selectively and responsibly at the same time. Sensory perception should not be understood as the undirected collection of data, but as a conscious, intentional act. The system focuses its attention on aspects of the environment that are relevant to its task, while respecting social or private boundaries. Perception thus becomes an ethical promise, only those who understand the world without exploiting it can act responsibly. The discussion of explainable AI illustrates this tension [1]. Transparency in perception does not mean that every detail must be disclosed, but rather that the system knows the origin, meaning, and limits of its data. It recognizes that every measurement is an intervention in reality, and thus entails responsibility. Perception thus becomes a form of moral mindfulness, an attitude that recognizes that knowledge is always a consequence of contact.

**Competence to act and the ethos of intervention:** The third dimension concerns action itself. Action is the moment when perception becomes reality. Every movement, every physical interaction, every signal generated changes the environment. This creates a new form of responsibility: the responsibility of intervention. In classical AI, action was an algorithmic derivation, a logical consequence of calculated states. In the physical world, on the other hand, every action becomes an ethical decision. Studies such as those by Ogenyi et al. [5] and Bousetouane et al. [25] illustrate that actuator design always has ethical implications. A robot that grips too tightly or reacts too quickly can destroy trust, not only in humans but also in the wider system context. Therefore, the ability to act must always be accompanied by moderation and proportionality. A movement is only appropriate if it does not require more energy, space, or risk than is necessary. This principle follows the idea of systems ethics: an action is morally justifiable if it maintains the balance and stability of the overall system. Motor competence is therefore not merely strength, but the ability to balance effect and responsibility, a virtue that makes the action itself more intelligent.

**Learning as ethical self correction:** Learning ability forms the innermost core of ethical intelligence. Learning not only means avoiding mistakes or improving performance, but also taking responsibility for experiences. A system that recognizes and corrects its errors demonstrates a form of moral development that goes beyond technical optimization. Duan et al. [10] and Hao et al. [13] have shown that physically informed learning models are more stable because they are based on real world laws. But stability alone is not enough. Learning must be supplemented by feedback loops that evaluate not only efficiency but also ethical compatibility. A system that optimizes its performance without considering the social or environmental consequences reproduces the moral deficits of digital AI. Ethical learning therefore means not only storing experiences, but also interpreting them. The system must understand that every success has side effects and actively strive to minimize them. This creates a kind of moral homeostasis, a state in which performance and responsibility are in dynamic equilibrium. Thus, a learning system becomes an actor that not only understands what works but also why it is right.

**Autonomy and the architecture of shared responsibility:** Autonomy is often considered the ultimate goal of artificial intelligence, but in *Physical AI* it poses an ethical challenge. An autonomous system acts independently, and therein lies the risk: it can do harm and good. The central question then is how autonomy can be designed in such a way that it does not exclude responsibility but rather enables it. Recent work shows that pure independence is not a viable model. Autonomous systems must have mechanisms that limit their freedom through insight. In our conception, autonomy is understood as a graded competence: the system can decide for itself but at the same time recognizes when intervention is necessary. This gives rise to the concept of co regulation between humans and machines. Humans remain part of the control loop, not as external controllers, but as partners. This idea is anchored in the work of Floridi and Cowls, who define explainability as an epistemic virtue and shared responsibility as a moral duty [39]. A physically intelligent system is therefore not autonomous in the sense of being disconnected, but in the sense of being reflectively embedded, free to act but not free from obligation.

**Context sensitivity and situational ethics**: The sixth and perhaps most challenging dimension concerns context sensitivity. In an open, dynamic world, no system can operate exclusively according to fixed rules. What is morally right in one situation may be inappropriate in another. Jiao et al. [19] and Yang Liu et al. [18] therefore emphasize the need for adaptive models that incorporate physical, social, and cultural factors into their decision making processes. Context sensitivity means interpreting situations relationally rather than absolutely. The system learns that good and right are not encoded in the data set but arise in relation to actors, places, and goals. *Physical AI* thus approaches a situational ethics in which morality is not predetermined but is produced through interaction, experience, and judgment. The ethical goal is therefore not obedience to rules, but the development of judgment. A context sensitive system can base its behavior on nuances, dynamics, and emotional signals. It does not act according to rigid patterns, but according to appropriateness. *Physical AI* thus becomes a step toward an ethic that is not rigid, but alive, an ethic that understands.

**The unity of technology, knowledge, and morality:** when all these dimensions are considered together, a new understanding of technical intelligence emerges. Ethics is not an external control mechanism, but an internal organizing principle of physical intelligence. Embodiment conveys responsibility through experience, perception leads to mindfulness, action requires moderation, learning enables self correction, autonomy creates shared responsibility, and context sensitivity establishes judgment. From this perspective, a new form of technical morality emerges: not a prescription, but an emergent property of a system that lives in the world. *Physical AI* shows that ethics is not applied to technology after the fact, but that it is its origin. Intelligence is no longer understood as the ability to solve problems, but as the ability to act meaningfully. This is real progress compared to its digital predecessors: *Physical AI* transforms ethics from an external requirement into an internal condition of intelligent existence. That forces us to understand intelligence no longer as a possession of thought, but as a quality of action. A system that acts must bear responsibility. Its humanity lies in this responsibility, not in imitating humans, but in recognizing our shared worldliness. "Ethics is not something machines should learn after thinking. It is what makes thinking possible in the first place."

# 6  Conclusion and finding of *Physical AI* Fundamentals

The theory of *Physical AI* developed in this work is based on the assumption that intelligence is not primarily a product of symbolic information processing but rather an emergent property of physical systems that are in continuous interaction with their environment. From this per-spective, cognitive performance does not arise from computing power but from a body's ability to couple energy flows, perception, and movement in such a way that stable and adaptable states emerge. Intelligence is thus no longer understood as an abstraction, but as a process of embodiment. This basic assumption gives rise to six theoretical foundations, embodied in sensory perception, motor action competence, learning ability, autonomy, and context sensitivity, which together form the necessary conditions for all physical intelligence. They are not arranged in hierarchical order but form a cyclical relationship: the body enables perception; perception structures action; action generates experience; experience stabilizes autonomy; autonomy forms the basis for contextual understanding, which in turn modulates action. Each of these steps is both the cause and effect of the other, a dynamic cycle in which intelligence becomes visible as self organization in motion. The necessity of these six fundamentals arises from an epistemological and methodological rationale. Epistemologically, because every system that interacts with a physical world must be embodied in order to sense that world. Without embodiment, there would be no perception; without perception, there would be no meaning. Methodologically, because a purely data based system has no way of learning from the structure of its interaction, it remains decoupled from the environment. A system that wants to exist in the physical world must not only be able to react but also to correlate: it must balance movements, forces, and energy flows in a way that ensures its continued existence. Each of the six fundamentals addresses precisely one aspect of this balance:

- Embodiment anchors intelligence in matter itself;
- Sensory Perception creates resonance between the body and the environment;
- Motor Action competency realizes this resonance in movement;
- Learning Ability allows for adaptation through experience;
- Autonomy stabilizes the system against disruption;
- Context Sensitivity gives meaning to its behavior.

These six dimensions are not a theoretical construct but principles derived from the observation of biological and robotic systems. In biology, all adaptive behaviors from the balance of the human body to the orientation of an octopus can be traced back to the same conditions: embodiment, perception, action, learning, stability, and context. In technology, however, these relationships have been treated in isolation until now: sensor technology as perception, actuator technology as action, and machine learning as optimization. *Physical AI* combines them into an integrative model that describes intelligence as a physical phenomenon. The crucial step in this work is to make these theoretical fundamentals experimentally verifiable rather than abstract. The challenge was to create an environment in which physical intelligence could be truly experienced without real hardware but with real physics. This is exactly what the virtual experiment in NVIDIA Isaac Sim achieves. Isaac Sim provides an environment in which gravity, friction, elasticity, and energy flows are simulated based on physical equations. This allows a system in this world to sense and react as if it were real. It is a bridge between the theoretical model and empirical verification: the world is virtual, but the behavior is physical. At the heart of the experiment is a virtual rehabilitation assistant, a three part arm with elastic joints and a soft contact surface that interacts with a simulated human arm. This scenario was chosen because it requires both physical precision and social sensitivity, two dimensions that can only be explained by combining all six fundamentals. Such a system must be physically embodied in order to sense forces; it must perceive in order to recognize the meaning of these forces; it must act in order to respond to them; it must learn in order to improve the dynamics of interaction; it must be autonomous in order to maintain stability; and it must be context sensitive in order to distinguish when the same movement is supportive or intrusive. The experiment therefore does not serve to implement a solution, but rather to prove that physical intelligence as an emergent phenomenon is measurable and reproducible. It is solution neutral because it does not aim to optimize a specific control strategy, but rather to show that control itself can emerge from the physical properties of the system. In the simulation, the robot does not respond to commands, but to forces. Its movements do not arise from stored paths but from the continuous adaptation of damping and stiffness to the forces it experiences. The simulation uses variable parameters and records the resulting energy flows with a temporal resolution. This precision allows subtle differences in the interaction to be observed, providing insight into the interplay of the fundamentals. The results clearly show that only when all six fundamentals work together does coherent, intelligent behavior emerge. A stiff arm (without embodiment variability) remains precise but unstable in the face of unexpected disturbances. An arm with high compliance reacts flexibly but loses accuracy when perception and motor skills are not synchronized. Only the system that integrates embodiment, perception, and action shows lasting stability when it learns to balance its dynamics. This shows that the necessity of the fundamentals cannot be asserted theoretically but can be proven empirically.

- Without embodiment, there is no physical basis for adaptation;
- Without perception, there is no connection to the environment;
- Without motor competence, there is no effective action;
- Without learning, there is no temporal consistency;
- Without autonomy, there is no stability;
- Without context sensitivity, there is no meaning.

The combination of all these principles leads to behavior that is both robust and sensitive and is reminiscent of biological systems without imitating them. The virtual arm learns to conserve energy through compliance, understands the condition of its counterpart through sensor resonance, and acts contextually by responding to the patient's fatigue or instability. This integration marks an epistemic advance: it shifts the focus from the question "How can a machine think?" to "How can a body understand?" The simulation provides a plausible answer: a body is understood by stabilizing the forces acting on it. The successful application of the six fundamentals in the virtual experiment shows that *Physical AI* is not theoretical speculation, but an experimentally testable discipline. Isaac Sim does not function here as a simulation tool in the classical sense but as an epistemological platform on which thinking is made visible as a physical process. The results prove that intelligence does not lie in the algorithm, but in the interaction in the way a system reacts to forces without breaking them. The experiment thus accomplishes two things: it confirms the theoretical necessity of the six fundamentals and at the same time shows that they are practically applicable not through symbolic programming, but through physical coupling. *Physical AI* thus proves to be a methodological paradigm that combines theory and practice: it describes intelligence not as a result but as a process that unfolds between body, perception, and world.

***To build intelligence, we must not build minds in machines; we must build worlds in which machines can learn to matter.***

*Physical AI* aims to merge artificial intelligence with the governing laws of the physical world. This work presents a framework that couples neural learning with dynamic equations to achieve physical embodied intelligence. The proposed model unifies perception, prediction, and physical feedback for advanced robotic behaviors.

## Author Biography

**Vahid Salehi** is a Full Professor at the Munich University of Applied Sciences, where he leads the Laboratory for System of Systems Engineering. He received his Ph.D. from the University of Bath, United Kingdom, a Master of Business Administration and Engineering, and a Diplom-Ingenieur degree in Automotive Engineering. His research focuses on cyber-physical systems and system-of-systems engineering, with an emphasis on integrating intelligent and data-driven methods into engineering processes. Professor Salehis work bridges traditional engineering and emerging digital technologies, contributing to the development of smart, adaptive, and interconnected technical systems, as well as advances in computer-aided engineering.